\documentclass[preprint,12pt]{elsarticle}
\usepackage[left=2.5cm, right=2.5cm, top=3cm, bottom=3cm]{geometry}
\usepackage{amsmath}

\usepackage{amssymb}

\usepackage{hyperref}
\hypersetup{
	colorlinks=true,
	linkcolor=blue
}
\usepackage{tikz}
\usetikzlibrary{shapes, arrows.meta, 3d}
\tikzstyle{arrow}=[draw, -latex]
\usepackage{subcaption}
\usepackage{todonotes}
\usepackage{layouts}
\usepackage{tikz-cd}
\usepackage{pgfplots}
\usepackage{subcaption}
\usepackage[linesnumbered,ruled,vlined]{algorithm2e}
\usepackage{float}
\journal{Engineering with Computers}
\usepackage{color,soul}

\begin{document}
\begin{frontmatter}
\title{Reducing Spatial Discretization Error on Coarse CFD Simulations Using an OpenFOAM-Embedded Deep Learning Framework}

\author[BCAM,UPV]{J. Gonzalez-Sieiro\corref{cor}}
\cortext[cor]{Corresponding author}
\ead{jgonzalez346@ikasle.ehu.eus}
\author[UPV,BCAM,IKERBASQUE]{D. Pardo}
\author[BCAM,TECNALIA]{V. Nava}
\author[CURTIN]{V.M. Calo}
\author[AACHEN]{M. Towara}

\affiliation[BCAM]{organization={BCAM - Basque Center for Applied Mathematics}, 
            city={Bilbao},
            country={Spain}}
        
\affiliation[UPV]{organization={University of Basque Country (UPV/EHU)},
	city={Leioa},
	country={Spain}}

\affiliation[IKERBASQUE]{organization={Ikerbasque - Basque Foundation for Sciences},
	city={Bilbao},
	country={Spain}}

\affiliation[TECNALIA]{organization={TECNALIA - Basque Research and Technology Alliance (BRTA)},
	city={Derio},
	country={Spain}}

\affiliation[CURTIN]{organization={School of Electrical Engineering, Computing and Mathematical Sciences, Curtin University},
	city={Perth},
	country={Australia}}

\affiliation[AACHEN]{organization={Software and Tools for Computational Engineering, RWTH Aachen
		University},
	city={Aachen},
	country={Germany}}

\begin{abstract}
We propose a method for reducing the spatial discretization error of coarse computational fluid dynamics (CFD) problems by enhancing the quality of low-resolution simulations using deep learning. We feed the model with fine-grid data after projecting it to the coarse-grid discretization. We substitute the default differencing scheme for the convection term by a feed-forward neural network that interpolates velocities from cell centers to face values to produce velocities that approximate the down-sampled fine-grid data well. The deep learning framework incorporates the open-source CFD code OpenFOAM, resulting in an end-to-end differentiable model. We automatically differentiate the CFD physics using a discrete adjoint code version. We present a fast communication method between TensorFlow (Python) and OpenFOAM (c++) that accelerates the training process. We applied the model to the flow past a square cylinder problem, reducing the error from 120\% to 25\% in the velocity for simulations inside the training distribution compared to the traditional solver using an x8 coarser mesh. For simulations outside the training distribution, the error reduction in the velocities was about 50\%. The training is affordable in terms of time and data samples since the architecture exploits the local features of the physics.
\end{abstract}

\begin{keyword}
Deep Learning \sep Computational Fluid Dynamics \sep  Finite Volume Method \sep OpenFOAM \sep  Discretization Error
\end{keyword}

\end{frontmatter}

\section{Introduction}\label{Intro}
Over the last decade, machine learning (ML) has arisen as a crucial technology for enhancing the solving of partial differential equations (PDEs)~\cite{ bruntonEDPs}. The field of fluid dynamics has been part of this revolution. Computational fluid dynamics (CFD) witnessed the introduction of novel learning techniques~\cite{ vin-brunton}. Non-linear PDEs describe fluid motion, usually showing chaotic spatiotemporal dependencies and structures with a wide range of scales, making Direct Numerical Simulation (DNS) unfeasible in most cases. More economical alternatives exist, like Large Eddy Simulations (LES)~\cite{ lesieur, Bazilevs2007}, in which only grid structures are solved while the rest are modeled, or Reynolds-Averaged Navier-Stokes (RANS)~\cite{ alfonsini}, in which all the scales are modeled. However, the resolution process can be complex and time-consuming, even with LES or RANS.

Enhancing CFD simulations through ML may accelerate the computations without deeply compromising accuracy or produce more accurate results with similar computational time. For this purpose, the ML model can completely substitute the traditional CFD solver, leading to reduced-order models (ROMs) or partially enhancing some specific components' performance. The ROMs exhibit accelerated time-to-solution but pay the cost of low fidelity and a narrow range of applicability~\cite{ taira, pant, Ghommem2014, iollo}. ROMs use linear data compression techniques like dynamic-mode decomposition (DMD)~\cite{ schmid} and principal-component analysis (PCA)~\cite{ jolliffe}, or non-linear compression like autoencoders~\cite{ lee}. When substituting a component of the traditional solver, the typical approach replaces the turbulent closure model ~\cite{ duraisamy}, which is one of the significant sources of error, with an ML model. Thus, Ling et al.~\cite{ling} outperform the traditional solvers using a deep neural network for RANS turbulence modeling. Alternatively, Obiols-Sales et al.~\cite{ obiols} speed up the convergence of RANS simulations by substituting the traditional solver with a subrogated model composed of a convolutional neural network (CNN) while the solution´s residual of the learned model is below a certain threshold; when this criterion is unsatisfied, they switch again to the traditional solver. Ajuria et al.~\cite{ ajuria} accelerate the algorithm by substituting the Jacobi solver with a CNN for the Poisson equation, one of the most expensive parts of the traditional numerical solver. 

A relevant point in ML computations is the generalization properties of the resulting approximation. Ideally, an ML model should enhance the simulation performance within multiple initial/boundary conditions, be accurate for unseen CFD simulations, and use an acceptable amount of data and time for training. Thus, Kim et al.~\cite{ jeon-kim-vin} seek to improve the precision of the Finite Volume Method (FVM)~\cite{ versteeg} combining predictions of a pre-trained neural network~\cite{ jeon-kim} and of a traditional solver. Another promising approach consists on reducing the discretization error on coarse meshes~\cite{ kochkov1}, applying a method called learned discretizations~\cite{ bar-sinai}. For example, Kochkov et al.~\cite{ kochkov2} estimate spatio-temporal operators on low-resolution grids through CNNs for accelerating DNS and LES simulations, learning from fine-resolution data. This technique calls the CFD solver during training in a solver-in-the-loop~\cite{ kiwon} ML model. This coupling demands an end-to-end differentiable framework; thus, translating the fluid solver to an automatic differentiable language is a stringent requirement~\cite{ baydin}. This limitation reduces the applicability of this kind of method to cases employing the already implemented physics models. Another drawback of the methodology is the demanding training process from a user point of view, because CNNs require a large dataset for predicting unsteady fluid flows~\cite{ leeyou}.

We extend the learned discretization technique to a real case with non-trivial geometries and more complex flow dynamics, such as the flow past a square cylinder problem~\cite{ trias}. In these convection-dominated flows, the discretization of the convective term is a critical source of error where the differencing scheme relating the cell´s face and center values plays a major role in stability and precision. Numerous scheme versions may be found in the literature depending on the conditions of the simulated case. We reduce the spatial discretization error on coarse meshes for unsteady RANS~\cite{ alfonsini} simulations by replacing the traditional differencing scheme with a feed-forward neural network, which learns from high-resolution data, which is previously restricted to the coarse grid discretization.

In contrast to previous works, we embed the open-source CFD code OpenFOAM~\cite{ jasak} in the training loop to exploit the vast range of physics and numerical solvers this program has implemented. Besides, our neural network's architecture exploits the local features of the flow to speed up the training process. We also develop a communication method between the ML framework and the CFD code to accelerate the training process compared with a hard-disk communication. Our algorithm applies the technique to our case study, but it is easily generalisable to any fluid case simulated in OpenFOAM. 

Our algorithm imitates a machine learning technique commonly applied in the image processing field called super-resolution~\cite{ yang}, following the idea of enhancing a low-quality image. Super-resolution consists of finding a mapping to reconstruct the high-resolution flow fields from the low-resolution data. Multiple methods have been proposed in the field of fluid dynamics~\cite{ fukami} varying the architecture, the flow conditions, and the loss function for imposing the physics. In contrast, our method learns a correction while keeping the coarse-grid discretization.  The method increases the accuracy of low-quality simulations online based on high-quality data, previously projected into the coarse discretization, employing data-driven models. Our model enforces physical constraints of the governing equations while enhancing certain parts of the traditional solver. Its main limitation resides in the necessity of data with an image format, that is, structured data generated from Cartesian meshes with cells of the same size.

The structure of the paper is as follows: Section~\ref{problem_description} describes the fluid problem and the different spatial discretizations involved. Section~\ref{baseline_solver} defines the physics to solve and explain the traditional solver's algorithm. Then, Section~\ref{deep_model} gathers the formal definition of the deep-learning model and the technical details of its implementation. Section~\ref{results} shows the numerical results of our approach and the performance comparison with the traditional solver. Finally, in Section~\ref{conclusions}, we summarise the findings and indicate the lines for future research. Additionally, ~\ref{generation_data} explains how we create the training data.

\section{Problem description}\label{problem_description}
\subsection{Fine discretization}
\begin{figure}[ht]
	\centering
	\includegraphics{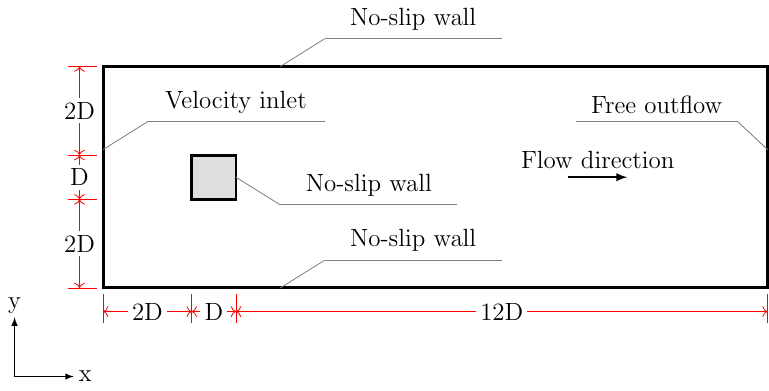}
	\caption{Schematic description of the computational domain and boundary conditions.}
	\label{fig:domain}
\end{figure}

The flow past a square cylinder is a well-known engineering problem with multiple industrial applications in offshore platforms~\cite{ yuliang}, bridge piers~\cite{ li}, high-rise buildings~\cite{ chen}, or coastal engineering~\cite{ gabi}. The state-of-the-art of this problem is well-developed with numerous numerical~\cite{ hongyi, clainche, bouris} and experimental studies~\cite{ luo, liu, dutta}, leading to a case that balances industry and research. There is no general convention between authors on the definition of the computational domain and boundary conditions; thus, we define our 2D problem (see Figure~\ref{fig:domain})  following~\cite{ xu}, where $D$ is the side of the square column.

According to the non-dimensional Reynolds number, which measures the ratio between inertial and viscous forces, we can identify different flow regimes~\cite{ xu, hongyi}. Most applications are interested in transition to turbulence or fully turbulent regimes, where attached vortex shedding appears in the wake of the obstacle and forms the von Kármán street vortex effect. These streets can be more or less stable depending on the turbulence grade of the flow, resulting in unsteady numerical simulations.

We fix the diameter of the square column $D=1$ m, an inlet velocity of 1 m/s, and a kinematic viscosity $\nu=2\cdot10^{-6}$ m/s$^2$, obtaining a Reynolds number $Re=5\cdot10^5$. Although this regime corresponds to a fully turbulent flow, we will obtain results closer to the laminar wake regime due to the use of unsteady RANS equations. In the case of simulating the flow in a high-fidelity manner, it would exhibit much finer length scales~\cite{ chatzi}. Nevertheless, building a mesh that is fine enough to describe the physics with fidelity is necessary. We define our fine spatial discretization $f$ as a structured mesh formed by 303,146 quadrilateral cells of the same size. The cell size of the mesh is selected to adequately recover the von Kármán effect, although at an increased computational cost. We establish a fixed temporal step of 0.005 s, obtaining a Courant-Friedrichs-Lewy (CFL) number~\cite{ CFL} equals 0.32. Although the simulated fine timestep is 0.005 s, we save the simulation result for each coarse timesteps (0.04 s) to force a time matching with the x8 coarse discretization. We simulate over 20 s and discard the first 4 s, obtaining 400 snapshots. We discard these steps to eliminate the simulation´s undesirable and unphysical behaviour prior to the wake formation.

From now on, we will present different operators and variables with the subscripts $c$ (coarse) or $f$ (fine), indicating to which spatial discretization we apply them. For the transformation of fine data into coarse dimensions, we introduce the projection operator $\mathcal{P}_f^c:\mathbb{R}^{m_f}\rightarrow\mathbb{R}^{m_c}$, which is applied to a generic scalar-valued field $\phi$ leading to: 
\begin{equation}\label{eq:pfc}
	\phi_{c,f} = \mathcal{P}_f^c(\phi_f) = \frac{1}{R}\sum_{j=1}^{R} \phi_{f}^i (j) \qquad\text{with $i \in \{1,...,m_c\}$},
\end{equation}
where the subscript $c,f$ denotes the fine data projected onto the coarse dimension, $R$ is the number of fine cells inside a coarse cell $i$, and $m_c$ is the number of cells of the destination coarse mesh. This operator runs through each cell of the coarse mesh and computes the arithmetic mean of the fine data contained within the considered cell. With this conservative averaging, we ensure that the resulting projected field fulfills continuity.

\subsection{Coarse discretization}
\begin{figure}[ht]
	\centering
	\begin{subfigure}[b]{0.48\textwidth}
		\centering
		\includegraphics[width=\textwidth]{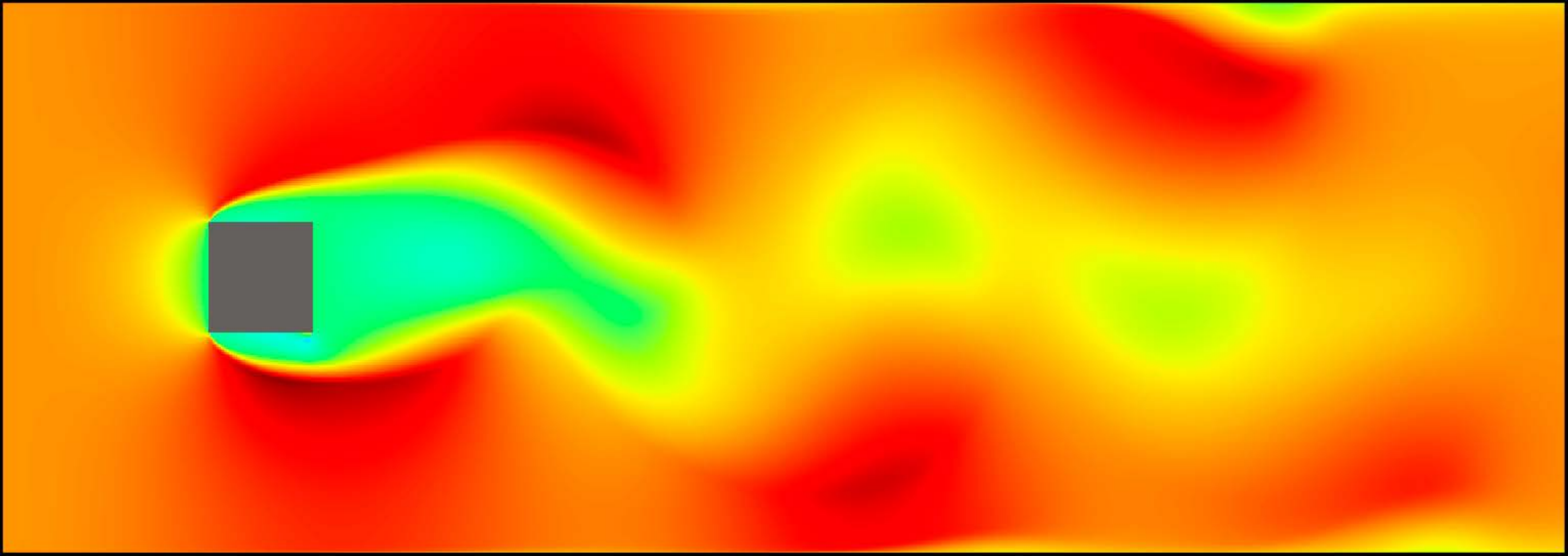}
		\caption{Result fine discretization.}
	\end{subfigure}
	\hfill
	\begin{subfigure}[b]{0.48\textwidth}
		\centering
		\includegraphics[width=\textwidth]{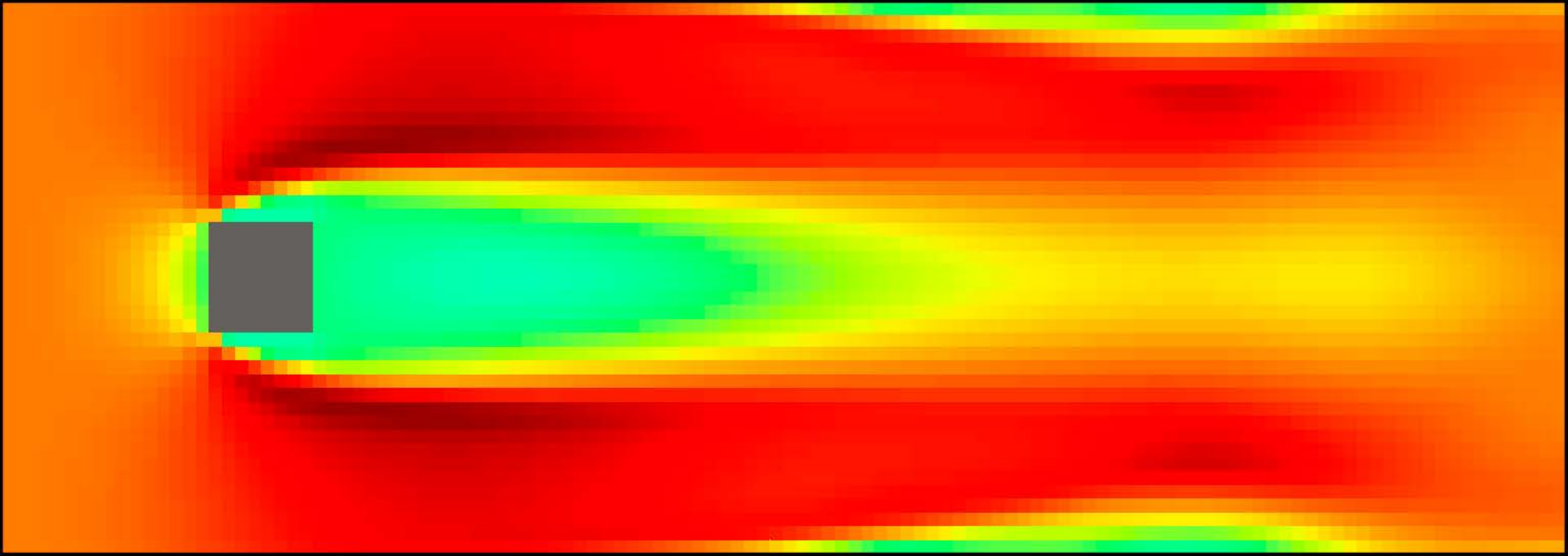}
		\caption{Result x8 coarse discretization.}
	\end{subfigure}
	\par\bigskip
	\begin{subfigure}[b]{0.48\textwidth}
		\centering
		\includegraphics[width=\textwidth]{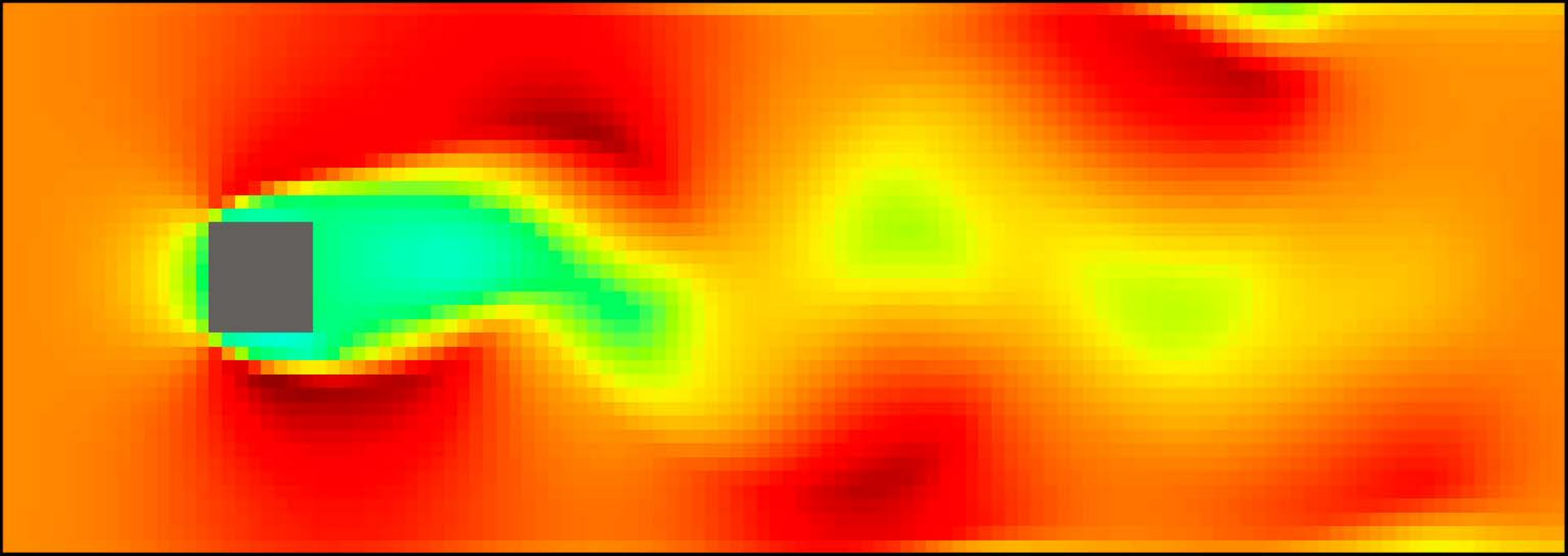}
		\caption{Down-sampled fine result.}
	\end{subfigure}
	\hfill
	\begin{subfigure}[b]{0.48\textwidth}
		\centering
		\includegraphics[width=\textwidth]{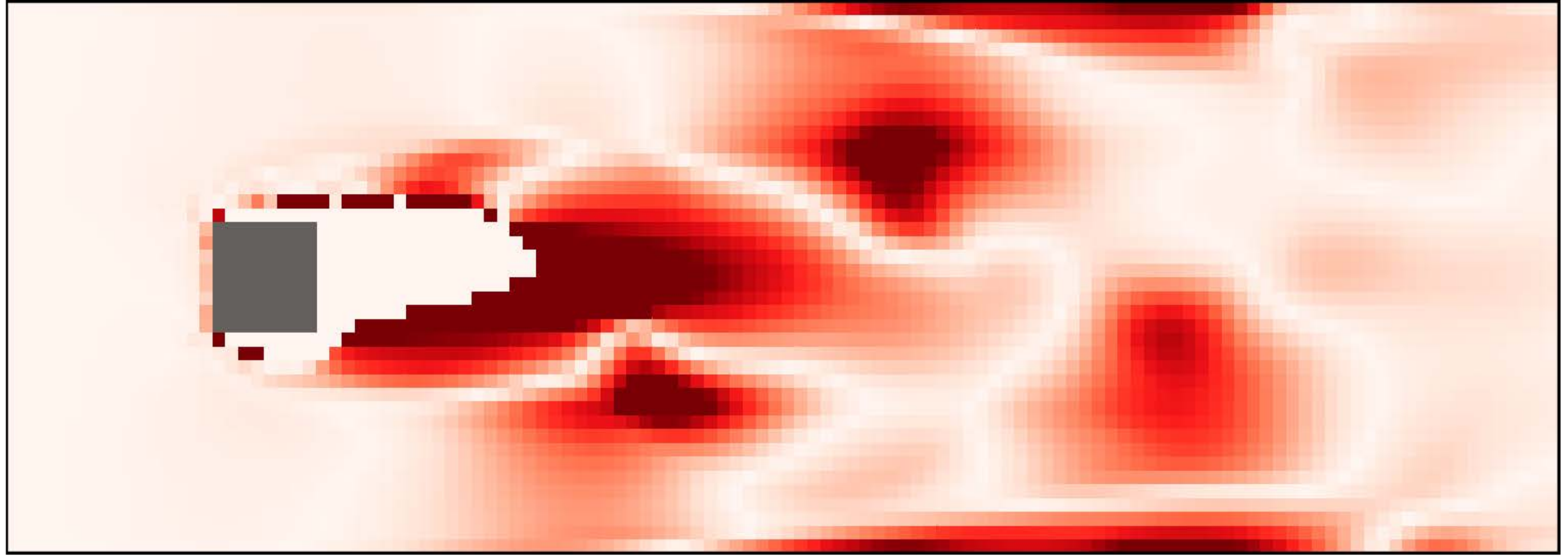}
		\caption{Error (\%) between (b) and (c)}
	\end{subfigure}
	\caption{Qualitative comparison of the results employing the different discretizations at $t=20$ s.}
	\label{fig:meshes}
\end{figure}

By down-sampling the fine mesh isotropically by a factor of x8, we define our coarse discretization $c$, which produces low-accuracy results. We apply our method to this mesh to enhance its solution from a learned model fed by highly-accurate data. We mantain the same CFL number of the fine case, that is 0.32; thus, coarse timestep is eight times bigger than the fine one. This mesh is composed by 4736 cells and 9296 interior faces. In order to illustrate the difference between both discretizations, we present Figure~\ref{fig:meshes}. It depicts a general comparison between the results obtained using both discretizations, the down-sampled fine result, and the error of the coarse simulation compared with the down-sampled case.

\subsection{Generalization cases}
\begin{figure}[ht]
	\centering
	\includegraphics{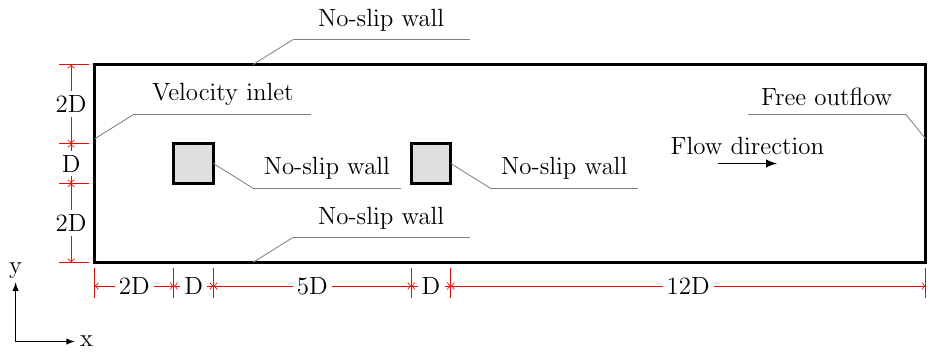}
	\caption{Schematic description of the computational domain and boundary conditions.}
	\label{fig:domain-tandem}
\end{figure}

We evaluate the generalization capabilities of our model testing its performance in two new cases derived from the one presented in Figure~\ref{fig:domain}. The first case consists of extending the simulation in time out of the training distribution. Thus, we let the model evolve from 20 s to 25 s, along 125 timesteps. The second experiment consists of extending the flow past a square cylinder in domain, inserting a second column and generating the two-columns-in-tandem case. The boundary and the flow conditions are maintained from the single-column case, as well as the size of the columns. Figure~\ref{fig:domain-tandem} describes this case.
\section{Baseline solver}\label{baseline_solver}
\subsection{Governing equations}
The system of PDEs describing the dynamics of an incompressible fluid flow is called the Navier-Stokes equations (NSE). We solve a Reynolds-averaged version of these equations, where all turbulence scales are modeled. Thus, the instantaneous values of the involved fields are decomposed into a mean component and a fluctuating one:
\begin{equation}\label{eq:Reynodls_decom}
\phi(\textbf{x},t)=\bar{\phi}(\textbf{x},t)+\phi'(\textbf{x},t).
\end{equation}
This simplification is commonly applied in industry due to the high computational cost of the alternative resolution techniques. The resulting equations are the Unsteady Reynolds-Averaged Navier-Stokes (URANS)~\cite{versteeg}:
\begin{equation}\label{eq:continuity}
\nabla\cdot\bar{\textbf{u}}=0,
\end{equation}
\begin{equation}\label{eq:momentum}
\dfrac{\partial{\bar{\textbf{u}}}}{\partial{t}}+
\nabla \cdot (\bar{\textbf{u}}\otimes\bar{\textbf{u}})-
\nu\nabla^{2}\bar{\textbf{u}}= 
-\dfrac{1}{\rho}\nabla{\bar{p}}+
\dfrac{1}{\rho}\nabla \cdot \tau^R,
\end{equation}
where $\bar{\textbf{u}}$ is the mean velocity field, $\bar{p}$ is the mean pressure field, the constants $\rho$ and $\nu$ are the density and the kinematic viscosity, and $\tau^R$ is the Reynolds stress tensor computed by the Boussinesq approximation. Due to this variable, a closure turbulence model is necessary. We use the $k-\omega$ SST~\cite{wilcox}, which adds two additional equations:
\begin{equation}\label{eq:k}
\dfrac{\partial{k}}{\partial{t}}+
\nabla \cdot (\bar{\textbf{u}}k)=
\tau^R:\nabla \bar{\textbf{u}}-
\beta^*k\omega+
\nabla\cdot[(\nu+\sigma^*\nu_t)\nabla k],
\end{equation}
\begin{equation}\label{eq:omega}
\dfrac{\partial{\omega}}{\partial{t}}+
\nabla \cdot (\bar{\textbf{u}}\omega)=
\alpha\frac{\omega}{k}\tau^R:\nabla \bar{\textbf{u}}-
\beta\omega^2+
\nabla\cdot[(\nu+\sigma\nu_t)\nabla \omega],
\end{equation}
where $k$ is the turbulent kinetic energy, $\omega$ is the turbulent specific dissipation rate, $\nu_t$ is the turbulent viscosity defined as the ratio between $k$ and $\omega$, and the rest of the variables are constants of the model. For the near-wall treatment of these turbulent variables, we employ wall functions to avoid the computational cost of resolving the boundary layer~\cite{ nieuwstadt}.

\subsection{Solver algorithm}\label{resolution_alg}
\begin{figure}[ht]
	\centering
	\includegraphics{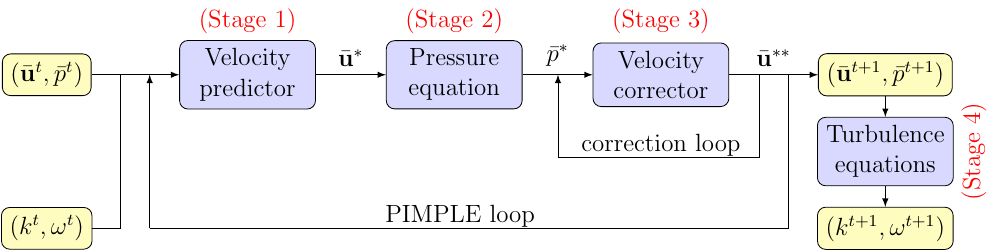}
	\caption{Schematic description of PIMPLE algorithm.}
	\label{fig:pimple}
\end{figure}

The complex PDE system describing the problem's physics requires an ad hoc iterative solving process. Multiple optimized algorithms exist depending on the fluid flow conditions; we use an operator-splitting method called Pressure IMplicit for Pressure-Linked Equations (PIMPLE~\cite{ versteeg}) for this case. This method is a predictor-corrector algorithm, formed by four main stages as described in Figure~\ref{fig:pimple}: (Stage 1) the velocity predictor, (Stage 2) the pressure equation stage, (Stage 3) the velocity corrector, and (Stage 4) the turbulence equations stage.

First, we construct a linearized momentum equation derived from~\eqref{eq:momentum} using the FVM and taking as input the variables at time $t$. The solution of this equation is the predicted mean velocity $\bar{\textbf{u}}^*$, which does not satisfy continuity~\eqref{eq:continuity}. During the second step, we form and solve a Poisson equation by combining~\eqref{eq:continuity}-\eqref{eq:momentum}, leading to a predicted mean pressure $\bar{p}^*$. During (Stage 3), the algorithm corrects the mean velocity based on the expected pressure computed in the previous step. This stage is repeated in the correction loop until the corrected mean velocity $\bar{\textbf{u}}^{**}$ is conservative. At the end of this stage, we obtain a candidate mean velocity and a candidate mean pressure, which can be sent back to (Stage 1) in case the convergence criteria are not satisfied. In that case, we would be restarting the PIMPLE loop taking $\bar{\textbf{u}}^{**}$ and $\bar{p}^*$ as the new input of the iteration. Once convergence is reached, we obtain the mean velocity and mean pressure for the time $t+1$. Finally, we jump into the final stage, where we solve the turbulence equations~\eqref{eq:k}-\eqref{eq:omega}, producing $k$ and $\omega$ at time $t+1$.

\subsection{Differencing schemes}\label{differencing}
The first step of the solver consists of linearizing the momentum equation to obtain a system of linear equations, which can be generalized for a cell $p$ as:
\begin{equation}\label{eq:linear_momentum}
a_p \bar{\textbf{u}}_p+\sum_{nb} a_{nb}\bar{\textbf{u}}_{nb} = \textbf{S}_p,
\end{equation}
where the sub-index $nb$ refers to the neighboring cells of $p$ and $\textbf{S}$ to the source. We obtain the $a_p$ and $a_{nb}$ coefficients by discretizing~\eqref{eq:momentum} using the Finite Volume Method (FVM), producing a matrix system, where $a_p$ and $a_{nb}$ are the diagonal and off-diagonal matrix coefficients, respectively. The mean velocity of each cell composes the vector of unknowns, and each cell's source term generates the system's right-hand side. Thus, for a 2D domain formed by $m$ cells, the matrix of coefficients will be a sparse $m\times m$ matrix with 5 non-zero entries per row and column since each cell will be only influenced by itself and its nearest neighboring cells.

In contrast to other numerical methods for solving PDEs, FVM works with fluxes throughout the faces of the cells, which makes the method especially convenient for fluid flows. However, it is necessary to use differencing schemes to relate face values with cell center values during this process. In case of velocities, we can define the mean face velocity $\bar{\textbf{u}}_F$ using a 2D general differencing scheme through the inverse operator $\mathcal{I}:\mathbb{R}^{2N}\rightarrow\mathbb{R}^{2}$, such that:
\begin{equation}\label{eq:diff_scheme}
\bar{\textbf{u}}_F = \mathcal{I}(\bar{\textbf{u}}_{(N)}) = \displaystyle\sum_{i=1}^{N}w_i\bar{\textbf{u}}_{i},
\end{equation}
where $N$ is the selected stencil around face $F$ and $w$ are the interpolation weights. The mean face velocity is not computed explicitly; in contrast, the interpolation weights $w_i$ contribute to the matrix of coefficients through $a_p$ and $a_{nb}$ of~\eqref{eq:linear_momentum}, following an implicit approach. A differencing scheme is an algebraic expression for relating a cell's face and center values based on some constant interpolation weights. Experts have developed numerous differencing schemes (e.g., Upwind, QUICK, TVD schemes~\cite{versteeg}) depending on the flow condition, resulting in variations of the stencil size and the interpolation weights. However, no general approach has been reached. For this reason, its choice is case-dependent, producing different results in accuracy and stability. Our approach builds a custom version of a differencing scheme employing deep learning.

\section{Deep-Convection Model}\label{deep_model}
\subsection{Formal description}\label{formal_descrip}
As mentioned in Section~\ref{problem_description}, our problem is fully turbulent, so the convective contribution is the most crucial term of~\eqref{eq:momentum}. The use of an appropriate mesh is essential for discretizing this term since a coarser mesh would considerably impact the accuracy due to its spatial discretization error. In particular, when we discretize the convective term using FVM, we obtain the following expression:
\begin{equation}\label{eq:discretisation}
\int_{V_p}\nabla\cdot(\bar{\textbf{u}}\otimes\bar{\textbf{u}}) \,dV
=\oint_{\partial{V}_P}(\bar{\textbf{u}}\otimes\bar{\textbf{u}}) \,d\textbf{S}
=\displaystyle\sum_{F}\bar{\textbf{u}}_F\otimes\bar{\textbf{u}}_{F}S_F
=\displaystyle\sum_{F}\bar{\textbf{u}}_{F}F_{F},
\end{equation}
where $\bar{\textbf{u}}_{F}$ is the mean velocity and $F_F$ is the mass flux, both at each face $F$ of the finite volume $P$. While the mass flux can be interpolated linearly from the cell center to the face, the mean face velocity must be interpolated carefully due to its major impact on the results~\cite{ ferziger}. In addition, a correct prediction of $\bar{\textbf{u}}$ in the momentum predictor step will produce an indirect improvement in the prediction of $\bar{p}$, $k$, and $\omega$.

As explained, the critical interpolation is carried out by a differencing scheme. Our approach substitutes the differencing scheme used for the convective term with a custom version based on data-driven discretizations~\cite{ bar-sinai}. Thus, we replace the constant interpolation weights $w_i$ of~\eqref{eq:diff_scheme} by a vector of optimized interpolation weights $\widetilde{\textbf{w}}_i$ generated by a feed-forward neural network. We can therefore define a modified inverse operator $\widetilde{\mathcal{I}}:\mathbb{R}^{3N}\rightarrow\mathbb{R}^{2}$, such that:
\begin{equation}\label{eq:mod_inverse}
\bar{\textbf{u}}_F = \widetilde{\mathcal{I}}(\bar{\textbf{u}}_{(N)}, \bar{p}_{(N)};\theta) = \displaystyle\sum_{i=1}^{N}\widetilde{\textbf{w}}_i\bar{\textbf{u}}_{i},
\end{equation}
where $\theta$ are the parameters of the neural network. In contrast to traditional differencing schemes, our approach takes the mean velocity and mean pressure as inputs and returns a vector of interpolation weights that are space and time-dependent. The resulting operators $\mathcal{I}$ and $\widetilde{\mathcal{I}}$ are local, meaning they compute the value at face $F$ based only on its neighborhood. However, we apply the local operators to all the internal faces of the domain, resulting in the mean face velocities of the domain $\bar{\textbf{u}}_{F(n)}$ as:
\begin{equation}\label{eq:gen_mod_inverse}
\bar{\textbf{u}}_{F(n)} := \widetilde{\mathcal{I}}(\bar{\textbf{u}}_{(N_j)}, \bar{p}_{(N_j)};\theta)= \displaystyle\sum_{i=1}^{N_j}\widetilde{\textbf{w}}_{ij}\bar{\textbf{u}}_{ij} \qquad\text{with $j \in \{1,...,n\}$},
\end{equation}
where $n$ is the number of faces of the domain. We encapsulate the rest of the resolution process described in Section~\ref{resolution_alg} in a forward operator $\mathcal{F}:\mathbb{R}^{5m+2n}\rightarrow\mathbb{R}^{2m}$, such that:
\begin{equation}\label{eq:forward}
\bar{\textbf{u}}^{t+1} := \mathcal{F} (\bar{\textbf{u}}_{F(n)}; \textbf{s}^t),
\end{equation}
where $\textbf{s}^t$ denotes the set of input variables:
\begin{equation}\label{eq:set}
\textbf{s}^t:=\bigcup^m_{j=1} \{\bar{\textbf{u}}^t_{(j)}, \bar{p}^t_{(j)}, k^t_{(j)}, \omega^t_{(j)}\},
\end{equation}
being $m$ the number of cells of the mesh. We fully define our model by composing these operators ($\mathcal{F} \circ \mathcal{I})(\textbf{s}^t)$ and applying them to a spatial discretization. In particular, we run the learned model on coarse meshes, which we indicate by the subscript $c$. We train it in such a way that the generated interpolation weights deliver a mean velocity $\bar{\textbf{u}}_c$ as close as possible to the mean velocity from the fine discretization projected onto the coarse one $\bar{\textbf{u}}_{c,f}$. Thus, we can formulate the training as the following minimization problem:
\begin{equation}\label{eq:min}
\theta^* :=\operatorname*{arg\,min}_\theta \sum_j \|(\mathcal{F}_c \circ \widetilde{\mathcal{I}}_c)(\mathcal{P}_f^c(\textbf{s}^t_f))- \mathcal{P}_f^c \circ (\mathcal{F}_f \circ \mathcal{I}_f) (\textbf{s}^t_f)\|,
\end{equation}
where $\mathcal{P}_f^c$ is the projection operator described in~\eqref{eq:pfc}, the subscript $f$ refers to the fine discretization and $\| \cdot \|$ is a discreate vector norm.

Our method, called Deep-Convection (DC), mimics the super-resolution techniques employed in image processing, which augment the accuracy of low-quality images based on high-quality data~\cite{ yang}. However, in our approach the enhancement is carried out maintaining the resolution and previously projecting the training data to the coarse-grid discretization. With this approach, the primary benefit is that it maintains the physical constraints of the governing equations while exploiting the benefits of a data-driven model for enriching certain parts of the traditional solver with high-resolution data.

\begin{figure}[ht]
	\centering
	\includegraphics{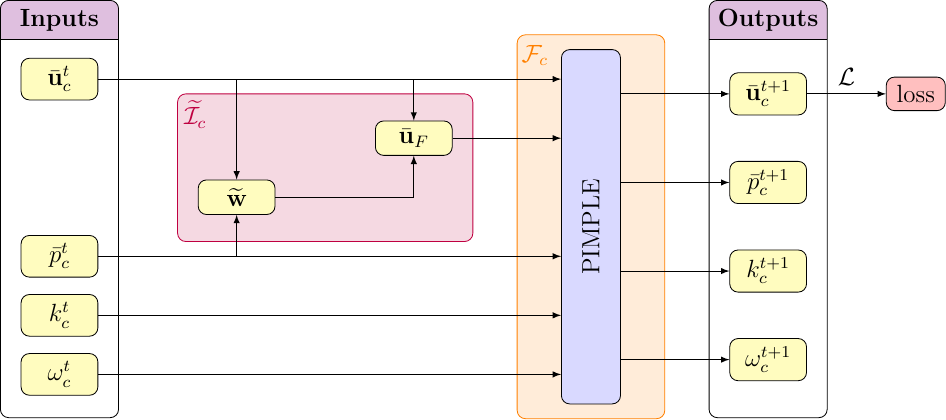}
	\caption{Schematic of the relation between variables and operators.}
	\label{fig:algorithm}
\end{figure}

The selected loss function to measure the error between the predicted values and the training data is a multi-step mean absolute percentage error, defined as:
\begin{equation}\label{eq:loss}
\Psi(\phi_c, \phi_{c,f}):=\frac{1}{B}\sum_{i=1}^{B}\frac{\sum_{j=1}^{T} \|\phi^{t_j}_{c,f}(i)-\phi^{t_j}_{c}(i)\|_{L^1}}{\sum_{j=1}^{T} \|\phi^{t_j}_{c,f}(i)\|_{L^1}} \cdot 100,
\end{equation}
where $\phi_c$ is a general scalar-valued field computed using the coarse discretization and $\phi_{c,f}$ is the projection of the fine data $\phi_f$ onto the coarse mesh $c$. In this case, $\phi_{c,f}$ is the training data or the ground truth. Due to its multi-step feature, we gather the solver prediction results over a number of accumulated timesteps $T$ and then compare them with the reference solution over this period. The number of accumulated timesteps $T$ is a crucial hyperparameter, which should be selected based on a trial-and-error procedure. A higher T will imply a better generalization of the model, but it will increase the training process cost. Different features of the problem, such as the complexity of the fluid flow, the number of interior faces of the domain, or the number of timesteps outside of the training distribution you pretend to simulate, will influence the value of T, so it is complicated to give an \textit{a priori} estimation. Since we work with mini-batches of training data, the summation over $B$ refers to evaluating all the cases composing the considered mini-batch. As we mentioned, we apply~\eqref{eq:loss} to the mean velocity, which is a vector field consisting of two components $\bar{\textbf{u}} = (\bar{u}^x, \bar{u}^y)$ in our case. Thus, we employ the previous equation with each component, resulting in the final loss as the evaluation of the following weighting loss function $\mathcal{L}$:
\begin{equation}\label{eq:weighting_loss}
\mathcal{L}(\bar{\textbf{u}}_c,\bar{\textbf{u}}_{c,f};\theta):= \kappa_x \Psi(\bar{u}^x_c,\bar{u}^x_{c,f}) + \kappa_y \Psi(\bar{u}_c^y,\bar{u}_{c,f}^y),
\end{equation}
where we set the weighting factors $\kappa_x$ and $\kappa_y$ based on each component's loss of the baseline coarse solver. We construct the loss function based just on the velocities for the sake of simplicity and with the intention of obtaining an indirect improvement on the rest of the variables~\cite{ kochkov2}. Including more fields in the loss function increases the training cost due to the reconstruction of additional intermediate derivatives, and generates a multi-objective optimization problem of coupled variables, commonly difficult to address~\cite{ sener}. Another possibility could have been to include the correction of other variables, adding them to the neural network's output while maintaining the loss function. This solution would also induce a more costly training process associated with a more complex optimization problem. A solution could be to introduce independent extra neural networks related to the correction of the rest of the variables and train them separately in a sequential way. However, exploring these alternatives is beyond the scope of the present paper, we may consider them in future works. Figure~\ref{fig:algorithm} outlines the algorithm and clarifies the relation between these operators. Thus, we start with the input variables at time $t$, then we apply sequentially the operators $\widetilde{\mathcal{I}_c}$ and $\mathcal{F}_c$, and finally obtain as output all the variables at time $t+1$. Then we apply $\mathcal{L}$ to $\bar{\textbf{u}}_c^{t+1}$ to get the corresponding loss.

Another difference compared with the traditional approach is that we are not taking the nearest-neighboring cells when building the stencil around a face. This means that the matrix of coefficients described in Section~\ref{differencing} is no longer sparse with 5 non-zero entries per row/column, which may induce stability problems during the iterative procedure for solving the system. In contrast, we follow a Deferred Correction~\cite{ moukalled} approach, commonly used for implementing high-order (HO) schemes. The technique consists of constructing the matrix of coefficients of~\eqref{eq:linear_momentum} using the Upwind differencing scheme while inserting on the r.h.s. of this equation the difference between the face velocity computed through the high-order scheme and the face velocity obtained using the Upwind scheme, computed explicitly. Thus, we define the convective flux using our model as:
\begin{equation}\label{eq:deferred}
F_F \bar{\textbf{u}}_F^{DC} = \underbrace{F_F \bar{\textbf{u}}_F^{U}}_{implicit} + \underbrace{F_F (\bar{\textbf{u}}_F^{DC} - \bar{\textbf{u}}_F^{U})}_{explicit},
\end{equation}
where the superscripts $DC$ and $U$ refer to the Deep-Convection and Upwind schemes.

\subsection{Deep learning architecture}\label{sec:architecture}
\begin{figure}[ht]
	\centering
	\includegraphics{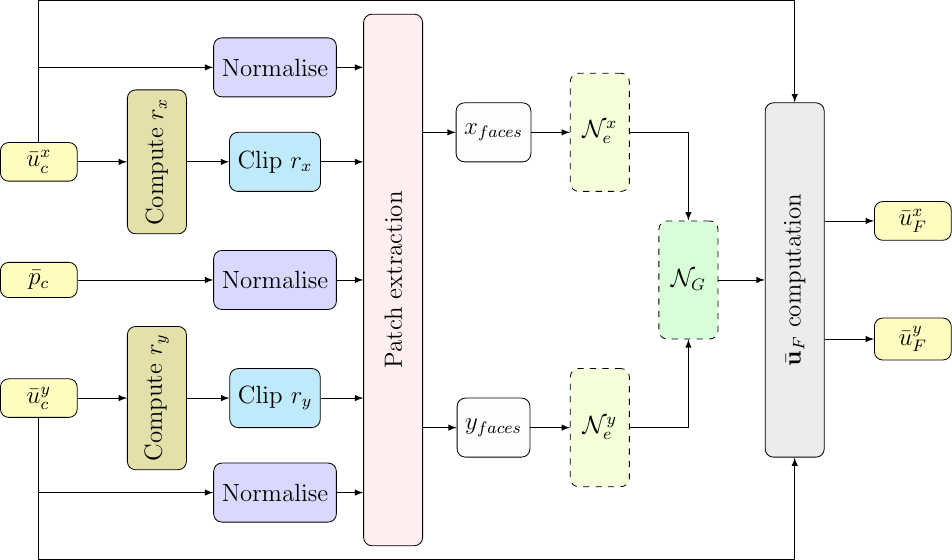}
	\caption{Architecture's sketch.}
	\label{fig:architecture}
\end{figure}

The architecture of the modified inverse operator $\widetilde{\mathcal{I}}$ comprises seven different types of operations, which we named as blocks, and we depicted their connectivities in Figure~\ref{fig:architecture}. We treat each velocity component independently, so we take three inputs at time $t$: each component of the mean velocity and the mean pressure. 

In the first block type, the $r$ variable for a general face $F$ of a cell $P$ is:
\begin{equation}\label{eq:r}
r_F:= \frac{\bar{u}_P-\bar{u}_U}{\bar{u}_D-\bar{u}_P},
\end{equation}
where $U$ and $D$ are the upwind and the downwind cell to $P$, respectively. The variable $r_F$ measures the ratio between the upstream and downstream gradients from face $F$. In this case,~\eqref{eq:r} assumes a 1D stream and a mesh with cells of the same size. Total Variation Diminishing schemes inspire the computation of this variable~\cite{ waterson}, commonly used due to their outstanding performance. The most important advantage of these schemes is that they preserve monotonicity, ensuring a stable and non-oscillatory solution. This feature is reached by monitoring the $r$ values, permitting the differencing scheme to act if the range of r is between $[0,2]$ and changing to the Upwind scheme otherwise. The $r$ variable can be interpreted as a second derivative or as a measure of the acceleration of the velocity field at $F$. Values below zero or higher than two mean a sharp change in the acceleration, being the face velocity totally influenced by the upwind or the downwind cell, respectively. For values between zero and two, our network should discover how to combine the neighboring cell center values. This interpretation justifies our second type of blocks, that clip the values of $r$ to the acting range $[0,2]$. The third block, following Figure~\ref{fig:architecture}, is the normalization one. Normalization of the neural network's inputs is a standard process that offers several benefits, like faster convergence or enhanced generalization. Moreover, it is also necessary in this case because the orders of magnitude of pressure and velocities are different. We normalize the velocities and the pressure in the following way:
\begin{align}
    \label{eq:norm_v}
\bar{u}^i_{norm}&:= \frac{\bar{u}^i}{max(\bar{u}^i)} \qquad\text{with $i \in \{x,y\}$},\\
\label{eq:norm_p}
\bar{p}_{norm}&:= \bar{p}-p_{atm},
\end{align}
where $i$ refers to each component of the mean velocity and $p_{atm}$ is the atmospheric pressure. Both types of normalization are zero preserving, which means the zero value in the normalized space is equal to the zero in the original one. In the case of the velocity, the normalized values are forced to be in the range [-1,1], preserving the original sign, which indicates the direction of the flow. The normalized value is equivalent to the pressure resulting from subtracting the reference pressure from the total pressure (summation of static and dynamic pressure). An extra transformation for enforcing $p_{norm}$ values between [-1,1] is unnecessary since the pressure variation for our flow conditions is very low. In cases with high-pressure variation, adding a transformation to the pressure similar to~\eqref{eq:norm_v} would be recommended.

\begin{figure}[ht]
	\centering
	\includegraphics{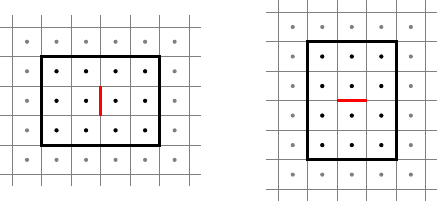}
	\caption{Description of the stencil used for vertical faces (left) and horizontal faces (right).}
	\label{fig:stencil}
\end{figure}

The next block, (see Figure~\ref{fig:architecture}), is patch extraction, which consists of going through every internal face of the domain and extracting the values of the neighboring cells after fixing the desired dimension of the stencil. We establish a stencil size of $4 \times 3$ for vertical faces and $3 \times 4$ for horizontal faces. Figure~\ref{fig:stencil} depicts a graphical description of the stencils employed. 

Special treatment is necessary for boundary patches or patches associated with faces where at least one of its neighboring cells falls outside the domain. In these cases, we fill the missed values with an artificial value, particularly with zeros, similarly to the padding applied in CNNs. This approach is linked to the election of the previous normalization since a cell with zero velocity and pressure in the normalized spaces means a wall cell with atmospheric pressure in the original spaces. We deal with the patches rather than the complete domain, aiming to exploit the local features of physics. CNNs were employed in previous works~\cite{ obiols,kochkov2} to extract the flow features accurately. However, our approach exploits the local features of the physics that describe the fluid dynamics to construct smaller neural networks, train the model with fewer data snapshots and reduce the training cost. Therefore, our procedure is “patch-based” rather than “domain-based”. Thus, we consider small patches, where the main benefits of CNNs are impaired. Hence, as we do not find significant reasons for selecting CNNs over other types of architecture, we decided to employ the simplest one, the fully connected neural networks. Other architectures will be explored in future works. The patch extraction block also separates the patches generated in the function of the internal face in $x_{\mathit{faces}}$ or vertical faces and $y_{\mathit{faces}}$ or horizontal ones. 

Afterward, the two groups of patches are sent, as Figure~\ref{fig:architecture} depicts, respectively to two neural networks $\mathcal{N}^x_e$ and $\mathcal{N}^y_e$ of the same shape, which act as encoders compressing the patch information ($r_x,r_y,\bar{u}^x,\bar{u}^y,\bar{p}$) to a latent space. We gather these networks' output and insert them into a third neural network $\mathcal{N}_G$, which generates the modified interpolation weights $\widetilde{\textbf{w}}$. Finally, in the last block, called $\textbf{u}_F$ computation, we obtain the mean face velocities based on the original mean velocities following~\eqref{eq:gen_mod_inverse}. Figure~\ref{fig:architecture} outlines the blocks defined and their connectivities.

The neural networks $\mathcal{N}^x_e$ and $\mathcal{N}^y_e$ are composed of four hidden layers of size 53, 49, 45, and 41, while the network $\mathcal{N}_G$ has two hidden layers of size 31 and 32. Altogether, the trainable parameters of the models add up to 22,036. In the case of the activation function, we employ the hyperbolic tangent to all the present layers. We initialize these neural networks following the standard Keras initialization, namely, the weights employ a uniform Xavier/Glorot initialization~\cite{glorot} and the bias are set to zero.

\subsection{Strong restrictions}
\subsubsection{Interpolation accuracy}
We enhance the numerical accuracy of the convective process using some modified interpolation weights $\widetilde{\textbf{w}}= (\widetilde{w}^{x},\widetilde{w}^{y})$ generated by our neural network; thus, we need to guarantee that the interpolation is at least first-order accurate like it is done for typical polynomial interpolation. We ensure the accuracy required by imposing the following restrictions:
\begin{equation}\label{eq:restr1}
\displaystyle\sum_{i=1}^{N}\widetilde{w}_{i}^x =1 \ \textnormal{and}\  \displaystyle\sum_{i=1}^{N}\widetilde{w}_{i}^y =1
\end{equation}
where $N$ is the stencil selected around face $F$. 

We impose this restriction following the technique described in~\cite{ bar-sinai, kochkov2}, which applies an affine transformation $\widetilde{\textbf{w}} = \textbf{A}\textbf{x}+\textbf{b}$ to the unconstrained output $\textbf{x}$ of the neural network. 
Starting from the linear system of equations $\textbf{C}\alpha=\delta$ constructed from the standard formula for generating the finite difference coefficients for a $N$-point stencil, we define the reduced constant matrix $\textbf{C}_r$ as the first $N-1$ rows of $\textbf{C}$. Thus, the matrix $\textbf{A}$ is the nullspace of the reduced constraint matrix, and the vector $\textbf{b}$ is an arbitrary valid set of interpolation weights. In this case, it equals the standard-finite difference solution $\alpha$. Thus, if we need $N$ interpolation weights, our neural network will generate $N-1$, and we obtain the remaining one using this affine transformation. We also explore other options, such as producing the $N$ unconstrained coefficients by the neural network and dividing each one by their sum. However, this alternative induced instabilities in the training process and produced initial coefficients further away from the optimum in the solution space. 

\subsubsection{Bounding neural network prediction on boundary patches}\label{sec:bounding}
In a learning model predicting temporal data, we distinguish three different regions: short-, mid-, and long-term simulations. The first region encompasses those simulations inside the training distribution, which the model should perform successfully. Mid-term simulations are those outside the training distribution but close to the seen data, so a well-designed model must predict them correctly. Long-term simulations are related to predictions far away from the training distribution, where a learning model just based on data is incapable of being accurate~\cite{jeon-kim-vin} due to the weak extrapolating capabilities of neural networks.

As mentioned in Section~\ref{sec:architecture}, a desirable property for a differencing scheme is monotonicity preserving, which means it does not create new undershoots and overshoots in the solution or accentuate existing extremes. TVD schemes achieve this feature through a flux limiter function, which changes to the Upwind scheme when the change in the velocity gradient is very sharp on both sides of the face. We observed that our neural network generally maintains this property, except for long-term simulations and boundary patches. The first case is a normal behavior in neural networks predicting time series related to the degradation of the solution after a great number of timesteps. This can be solved in our case by extending the training to a higher value of accumulated timesteps $T$. The second issue is not a desirable behaviour and must be prevented to avoid an early degradation of the solution.

The the neural network can produce unphysical predictions for some boundary patches, especially around the column. This phenomenon is related to the near-wall treatment applied to the turbulent quantities and the size of the cells of the first wall layer. Consequently, the wall function damps the solution on these cells, producing accurate results for short-term simulations even if the neural network prediction is completely wrong. However, this small error accumulates and arises during mid-term simulations, producing a wrong generalization. Thus, we implement a 1D limiter to bound the prediction of our neural network in these types of patches, inspired by the limiters of TVD schemes. Thus, considering a face $F$ surrounded by its owner cell $o$ and its neighbor cell $n$, we define the lower $b_-$ and upper $b_+$ bounds of our limiting criteria, such that:

\begin{equation}\label{eq:bounding1}
\begin{aligned}
&b_- = min(\bar{u}_o,\bar{u}_n)-\lambda |min(\bar{u}_o,\bar{u}_n)|, \\ 
&b_+ = max(\bar{u}_o,\bar{u}_n)+\lambda |max(\bar{u}_o,\bar{u}_n)|,
\end{aligned}
\end{equation}
where $\lambda$ is a constant for extending or narrowing the range of action of the limiter. We check the mean face velocities predicted in the boundary patches by our neural network so that if they are out of the bounds in~\eqref{eq:bounding1}, we change to the Upwind scheme. Hence, we obtain our bounded mean face velocity $\bar{u}^*_F$ as:
\begin{equation}\label{eq:bounding2}
\bar{u}^*_F:=\left\{
\begin{aligned}
Upwind(\bar{u}_o,\bar{u}_n) \qquad &if\  \bar{u}_F < b_-, \\ 
\bar{u}_F \qquad &if\ b_- \leq \bar{u}_F \leq b_+, \\
Upwind(\bar{u}_o,\bar{u}_n) \qquad &if\ \bar{u}_F > b_+, \\ 
\end{aligned}
\right .
\end{equation}
being $\bar{u}_F$ the unconstrained velocity generated by the model. As we treat each velocity component uncoupled, this limiter is applied independently to each component.

We apply this constraint in the prediction stage once the model is trained in an unconstrained way. Otherwise, the generalization in the surroundings of the column deteriorates. A correct selection of $\lambda$ is crucial since, in case of underestimating or overestimating this parameter, we will not be correcting enough the solution or spoiling the neural network's prediction, respectively. We set $\lambda=0.3$ experimentally using the training data once the neural network is completely trained. Besides, we employ the same mesh description as OpenFOAM, so we can easily filter and detect boundary patches.
 
\subsection{Training process}\label{training_process}
We built the dataset for training the model from a CFD simulation of the domain described in Section~\ref{problem_description} simulated for 20 s. However, during the first 4 s, the flow shows a transient and undesirable behavior until the wake starts to be formed. For this reason, we discard the first 100 timesteps, resulting in 400 training snapshots. We create the input-output entries of the dataset following~\ref{generation_data}, after projecting the data to the coarse dimension and after fixing the number of timesteps accumulated $T$, according to the multi-step loss function~\eqref{eq:loss}. Due to the center-to-face approach, we generate a number of samples equal to the number of internal faces from a snapshot of the complete domain. The number of epochs and training samples depends directly on the number of timesteps accumulated, which will be defined for each case in the results section. Algorithm~\ref{al:training} formally describes the multi-step training process.

We employ a batch size of 30 and a stochastic gradient descent optimizer, Adam~\cite{adam}, with an initial learning rate of $\eta^0=0.01$ and the rest of the parameters as default. Moreover, we use a callback to reduce the learning rate on the fly if the training loss did not decrease after a certain number $Z$ of epochs. We monitor the training convergence by a stopping criterion that checks if the loss function plateaus during a certain number of iterations (around 20), once we reach the lowest learning rate we set in the callback.

\begin{center}
\begin{algorithm}[H]\label{al:training}
	\SetKwInput{KwInput}{Input}
	\SetKwInput{KwOutput}{Output }
	\SetKwFunction{Build}{BuildTrainingData}
	\SetKwFunction{Dec}{Decrease}
	\caption{Multi-step training process}
	\KwInput{$\theta^0$,$\eta^0$,$T$,data\tcp*{Initial variables and training data}}
	\KwOutput{$\theta^*$,$\ell^*$\tcp*{Optimal weights and optimal loss}}
	
	$\theta^0_t \gets \theta^0$ \;
	\For{$t\leftarrow1$ \KwTo $T$}{
		$(\textbf{s}_{c,f}^t$, $\bar{\textbf{u}}_{c,f}^t) \gets$ \Build{data,$t$}\tcp*{From CFD simulation}
		$\ell_t^0 \gets \mathcal{L}((\mathcal{F}_c \circ \widetilde{\mathcal{I}}_c)(\textbf{s}_{c,f}^t), \bar{\textbf{u}}_{c,f}^t;\theta^0_t)$; $\theta_t^1 \gets \theta_t^{0} - \eta^0 \frac{\partial \ell }{\partial \theta}(\theta_t^0)$\; $\theta^*_t \gets \theta_t^0$; $\ell^*_t \gets \ell_t^0$; $i \gets 1$\;
		\While{not converged}{
			$\ell_t^i \gets \mathcal{L}((\mathcal{F}_c \circ \widetilde{\mathcal{I}}_c)(\textbf{s}_{c,f}^t), \bar{\textbf{u}}_{c,f}^t;\theta_t^i)$; $\theta_t^{i+1} \gets \theta_t^i - \eta^i_t \frac{\partial \ell }{\partial \theta}(\theta_t^i)$\;
			
			\If{$\ell^i_t > \ell^*_t$ during more than Z iterations}
			{
				$\eta^{i+1}_t \gets$ \Dec{$\eta^i_t$}; $\theta^{i+1}_t \gets \theta^*_t$
			}
			\Else
			{
				$\theta^*_t \gets \theta^i_t$;
				$\eta^{i+1}_t \gets \eta^i_t$;
				$\ell^*_t \gets \ell^i_t$; 
			}
			$\ell^* \gets \ell^*_t$; $\theta^* \gets \theta^*_t$; $i \gets i+1$
		}
		$\theta^0_{t+1} \gets \theta^*$;
	}
	\Return $\theta^*$,$\ell^*$
\end{algorithm}
\end{center}

\subsection{Implementation details}
The fluid dynamic part of the method has been implemented using OpenFOAM-v2112~\cite{ jasak}, while the deep learning framework has been TensorFlow~\cite{ tensorflow} in Python 3.10. Although TensorFlow has a C++ API, we decided to implement the data-driven part in Python to easily create and deploy learning models. Besides, the machine learning community has widely developed models using Python. Hence, another reason for this decision is not to limit the use of the model and expand the outreach to a broader community. The training and the simulations were performed in a workstation with an Intel(R) Xeon(R) W-2295 CPU equipped with 188 GB of RAM.

\subsubsection{Communication between programs}
\begin{figure}[ht]
	\centering
	\includegraphics{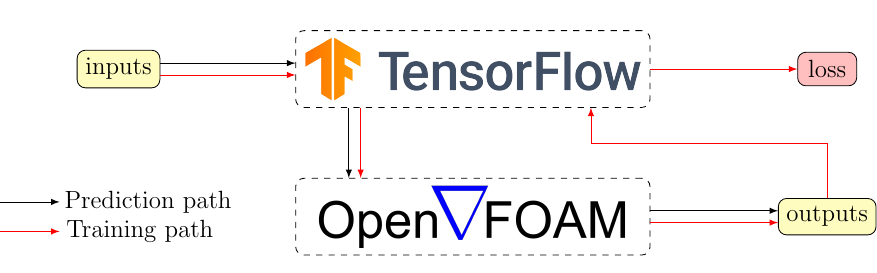}
	\caption{Dataflow between software.}
	\label{fig:dataflow}
\end{figure}

We work with two different pieces of software that must communicate during prediction and training. Moreover, they are written in different programming languages, making communication more complex. The critical step becomes during training since this process requires bidirectional communication, as Figure~\ref{fig:dataflow} depicts. The training process also requires fast communication because the dataflow path is repeated hundreds of thousands of times while optimizing the network parameters. 

We found libraries in the literature that permit the execution of Python codes from OpenFOAM, like PythonFOAM~\cite{ pythonfoam} or Pybind11~\cite{ pybind11}, as well as codes that launch OpenFOAM simulations from Python, like PyFOAM~\cite{ pyfoam}. However, none of these allow communication with the necessary characteristics. For this reason, we developed a new, fast, and bidirectional communication method. We based our method on RAM communication using the Python binding Ctypes~\cite{ ctypes} for transferring data between C++ and Python and the Shared Memory technique~\cite{ shm} to allocate data in RAM. Furthermore, we created a library for reading/writing data in OpenFOAM ASCII input/output format from Python, converting it to Python data and vice versa. For testing, we implemented a hard-disk communication to compare the performance of our method. We observed an 80\% run-time reduction in the training process using the in-RAM method compared to the basic hard disk one.
 
It is worth mentioning that TensorFlow divides the entire dataset into mini-batches of data to exploit GPU parallelization. In our case, we ran both the TensorFlow and the OpenFOAM parts in CPU, but we parallelized the execution of the simulations that compose each batch using the OpenMP~\cite{ openmp} package. That is why our batch size is not a power of two as usual since we are limited to the number of CPU threads. In the next version of the code, we plan to make the deep learning part fully GPU-compatible, so that the execution of the code becomes a sequential GPU-CPU-GPU procedure.

\subsubsection{Differentiation of OpenFOAM}\label{sec:differentiation}
In search for the optimal weights of the neural network, TensorFlow applies an algorithm called backpropagation~\cite{ goodfellow} based on automatic differentiation (also known as algorithmic differentiation, or just AD)~\cite{ griewank}. This technique assumes that any differentiable function $F:\mathbb{R}^n\rightarrow\mathbb{R}^m$ can be described as the composition of elementary functions for which the exact derivative is known. Then, the derivative of the loss w.r.t. the network parameters is obtained by applying the chain rule on this composition. Once this derivative is computed, minimizing the loss using an optimizer is possible. This method is feasible because the framework of TensorFlow is fully differentiable, so any code can be backpropagated as long as it is written using the package's functions.

\begin{figure}[ht]
	\centering
	\includegraphics{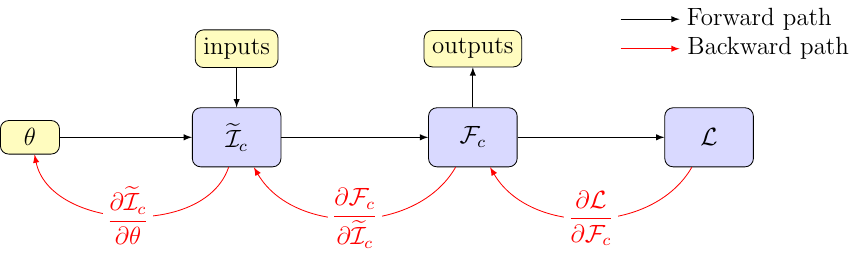}
	\caption{Representation of the forward and backward directions of the algorithm.}
	\label{fig:differentiation}
\end{figure}

In Section~\ref{formal_descrip}, we describe our method as a composition of the functions $\mathcal{I}_c$ and $\mathcal{F}_c$, while the operator $\mathcal{L}$ applied to this composition returns the loss, as Figure~\ref{fig:differentiation} describes. We obtain the optimum parameters of the neural network by computing the derivative of the loss w.r.t. the parameters of the network $\theta$, which in our case is defined as:
\begin{equation}\label{eq:chain_rule}
\frac{\partial \mathcal{L}}{\partial \theta}=\frac{\partial \mathcal{L}}{\partial \mathcal{F}_c} \cdot \frac{\partial \mathcal{F}_c}{\partial  \widetilde{\mathcal{I}}_c}\cdot \frac{\partial \widetilde{\mathcal{I}}_c}{\partial \theta}.
\end{equation}
The terms $\partial \mathcal{L} / \partial\mathcal{F}_c$ and $\partial \widetilde{\mathcal{I}}_c /\partial \theta$ are automatically calculated by TensorFlow through AD, but the derivative $\partial \mathcal{F}_c /\partial \widetilde{\mathcal{I}}_c$ must be specified by the user. This is because the operator $\mathcal{F}_c$ is carried out by OpenFOAM, which is implemented in the non-differentiable C++ programming language. Thus, we need to compute or approximate the derivative of the forward operator, which in terms of variables is equivalent to $\partial \bar{\textbf{u}}^{t+1}_c /\partial \bar{\textbf{u}}_F$.

We fill this gap using a discrete adjoint version of OpenFOAM developed by Towara~\cite{ towara}. This method applies AD by operator overloading to the program's core instead of solving the adjoint equations as a continuous adjoint software. By evaluating the chain rule on a per-operation level through the whole solver execution, it calculates the derivative we need. Additionally, the code allows to efficiently evaluate Jacobian-vector (tangent/forward AD) and vector-Jacobian (reverse/adjoint AD) products, leading to an economical and adequate solution for our problem.
Note the strong relation in concepts between adjoint AD and backpropagation in ML~\cite{ baydin}.

Other alternatives, such as reconstructing the missed derivative using finite differences or employing a Bayesian optimizer, are computationally expensive, making these options unfeasible for our purpose.

\section{Results and Discussion}\label{results}
\subsection{Sensitivity study to the reduction of mesh resolution}
We first analyze the model's behavior induced by the resolution reduction from the fine to the coarse mesh. For this purpose, we define the reduction factor $F_r$ as the ratio between the number of cells of the fine mesh and the coarse one in each of the main directions:
\begin{equation}\label{eq:reduction}
F_r=\frac{m_f^x}{m_c^x}=\frac{m_f^y}{m_c^y},
\end{equation}
where $m$ is the number of cells. As the reduction between meshes is isotropic, the reduction factor is the same in the directions of $x$ and $y$. We conducted the sensitivity study varying the resolution factor and observing the error by the multi-step mean absolute percentage error $\Psi$ under the same training conditions. The training with each mesh predicts just one timestep, so we fix the number of accumulated time steps $T$ to 1, obtaining 399 training snapshots. We establish the number of epochs or training iterations to 100, with a callback to reduce the learning rate by a factor of 0.5 if the loss does not decrease after six epochs. The obtained weighting factors (described in~\eqref{eq:weighting_loss})  were almost constant for each of the coarse discretizations analysed. For this reason and for carrying out a fair comparison, we employed the same weighting factors in this sensitivity analysis.

\begin{figure}[H]
	\centering
	\includegraphics{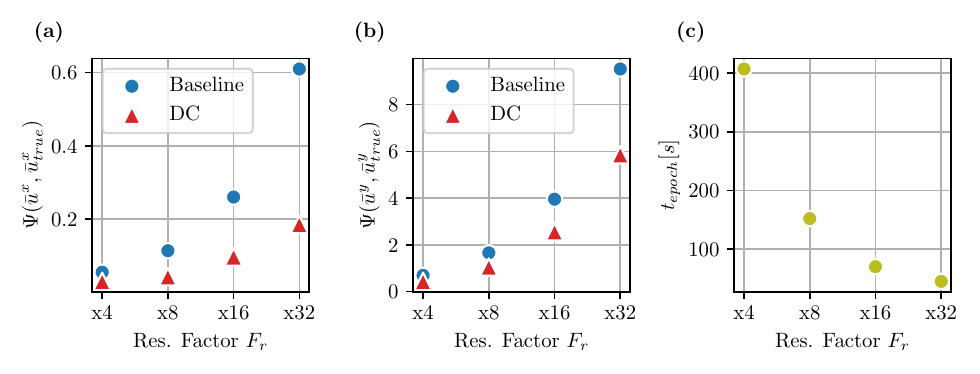}
	\caption{Results of the sensitivity analysis for reducing the mesh resolution. (a) Error on the x-component of the velocity. (b) Error on the y-component of the velocity. (c) Training computation time per epoch.}
	\label{fig:resolution}
\end{figure}

Figures~\ref{fig:resolution}(a) and~\ref{fig:resolution}(b) gather the resulting value of the loss function of each component of the velocity once the training process is completed (last epoch value). Regarding the effect of the resolution factor, we observe a lower degradation in the prediction of the Deep-Convection model compared with the baseline solver for the x-component of the velocity. A similar degradation is observed for the y-component. For a resolution factor of $\times 4$, the reduction in the loss for the $\bar{u}_x$ is 45\% and for $\bar{u}_y$ 37\%, while in the case of $\times 32$, the x-component exhibits a reduction of 70\% and the y-component 39\%. These percentages may be controlled by varying the weighting factors of~\eqref{eq:weighting_loss}. 

Figure~\ref{fig:resolution}(c) shows the computational time required to execute one epoch of the training $t_{epoch}$. We observe that this computation time decreases drastically with the increase of the resolution factor. This makes the training for a reduction of x2 unfeasible due to its computational cost and the feasibility of the x4 case dependent on the available resources. For this reason, to test the model in the worst conditions, we decided to continue the experiments with the mesh with an x8 reduction factor.

\begin{figure}[H]
	\centering
	\includegraphics{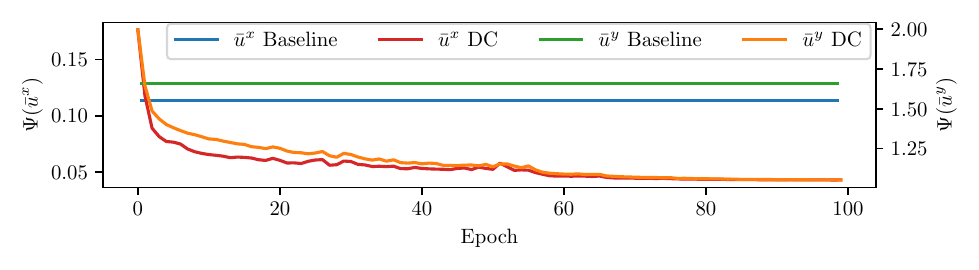}
	\caption{Training curves and loss of the baseline solver for both velocity components for an x8 mesh.}
	\label{fig:training_curves}
\end{figure}

One advantageous feature of the model is the fast convergence shown during training. Figure~\ref{fig:training_curves} shows how the Deep-Convection model outperforms the baseline solver after a couple of epochs for both velocity components. Moreover, it reaches 81\% and 71\% of the total loss reduction for x and y components within the first ten iterations. In contrast to previous  “domain-based” models presented in Section~\ref{Intro}, which usually need over 1000 epochs to complete the training, our “patch-based” model achieves convergence in around 70 epochs. This feature is vital when training with more than one accumulated timestep because it permits a reduction in the training duration. The rest of the resolution cases show a similar training curve, with almost identical speed of convergence. Regarding Figure~\ref{fig:training_curves}, we also observe the good behavior of the minimization algorithm, producing a smooth reduction in the loss. Thus, we can state the correct reconstruction of the missed derivative, described in Section~\ref{sec:differentiation}, and the successful coupling between TensorFlow and OpenFOAM.

\subsection{Performance on mid-term simulation}
The promising results of the previous section show that the model can predict a single timestep effectively. However, it is necessary to evaluate the mid-term performance of the model since stability and a correct generalization are crucial for obtaining a valuable method. Thus, we trained the model from 1 to 6 accumulated timesteps until convergence and from 7 to 13 with a fixed number of epochs. Generalization properties can be improved by training with a large value of T. Unfortunately, as T increases, the training costs also increase, so it is necessary to find a balance between generalization and cost. In addition, the increase in T must be progressive. Otherwise, the training will diverge. In the first stages of the training with $1\leq T\leq 6$, we can execute as many epochs as we want since the cost is affordable, so we extend the execution until convergence. For $T \geq 6$ the execution costs increase. However, by executing a limited number of epochs (without reaching full convergence), the method retains adequate generalization properties and accuracy since the highest loss decrease occurs within the first training iterations (see Figure~\ref{fig:training_curves}). We also employed the callback mentioned in the previous section to reduce the learning rate on the fly.

\begin{figure}[H]
	\centering
	\includegraphics{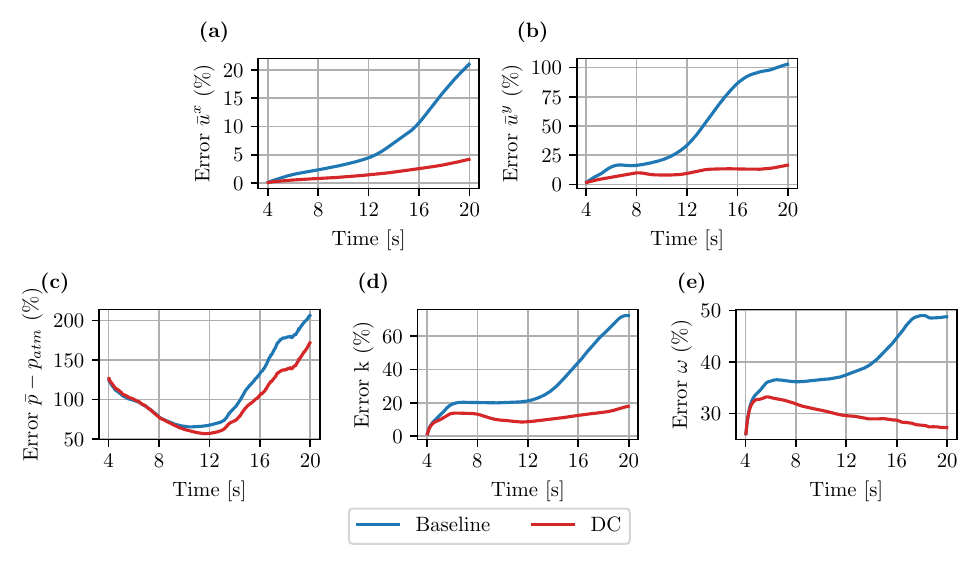}
	\caption{Comparison of results for a mid-term simulation between the baseline solver and the Deep-Convection. (a) Error on the x-component of the velocity. (b) Error on the y-component of the velocity. (c) Error on the pressure. (d) Error on the $k$ variable. (e) Error on the $\omega$ variable.}
	\label{fig:mid_term_all}
\end{figure}

The experiment simulates of the training data from the first to the last. Thus, the baseline solver and the Deep-Convection start from the same initial conditions at 4 s and evolve until reaching 20 s, going along 399 timesteps. It is evident that the longer the simulation, the higher the effect of the coarse mesh in the error since the error accumulates from one timestep to another. Furthermore, the relative error in the y-component of the velocity is higher than in the x-component due to the manifestation of the von Kármán effect as an oscillation along the y-axis. Figure~\ref{fig:mid_term_all} gathers the evolution of all the variables involved in the simulation.

\begin{figure}[H]
	\centering
	\includegraphics{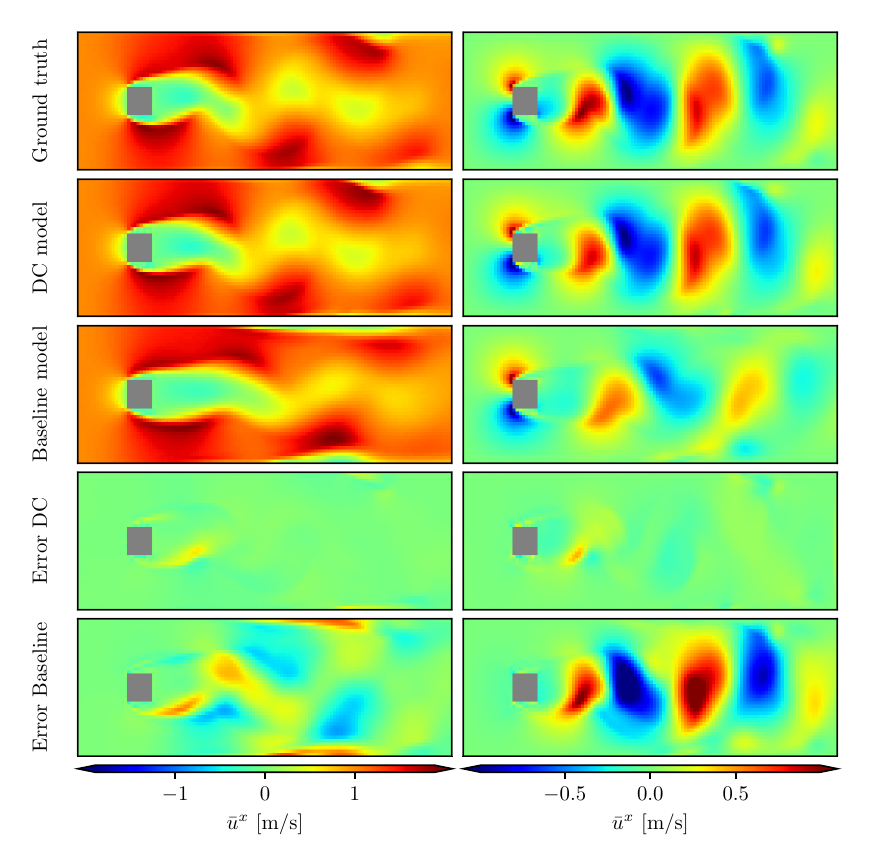}
	\caption{Comparison of each component of the velocity field and error at $t=20$ s between the ground truth, the Deep-Convection model, and the baseline solver.}
	\label{fig:mid_term_last_u}
\end{figure}

Figures~\ref{fig:mid_term_all}(a) and~\ref{fig:mid_term_all}(b) show that the Deep-Convection model reduces the error more than 50\% both in the x- and y-velocity components until $t=12$ s. The reduction is even higher for the following timesteps, reaching a reduction above 75\% of both velocity components at $t=20$ s. This improvement in the performance of the DC model after $t=12$ s is due to the full development of the flow at this point, being the under-developed flow harder to enhance. The relative error in the y-component of the velocity is almost five times higher than in the x-component, as anticipated, and the increase of the error with time is much more controlled with the DC model than with the baseline one. At $t=20$ s, the velocity results are completely different, being the ones produced with the DC model more similar to the ground truth, as depicted in Figure~\ref{fig:mid_term_last_u}.

Figures~\ref{fig:mid_term_all}(d) and~\ref{fig:mid_term_all}(e) show a similar behaviour of the turbulent quantities $k$ and $\omega$, reducing the error from 70\% to 20\% and from 50\% to 25\% at $t=20$ s, respectively. Like the velocity, the error reduction increases after $t=12$ s, when the flow becomes a fully developed flow. Hence, the turbulent variables modeling improves indirectly by enhancing the velocity with fine-scale effects. Figure~\ref{fig:mid_term_all}(c) depicts that the pressure also reduces its error, but the reduction is far less significant. Before $t=12$ s, the pressure error of the DC model is practically the same as the baseline one, increasing after this point until around 20\%.

\begin{figure}[ht]
	\centering
\includegraphics{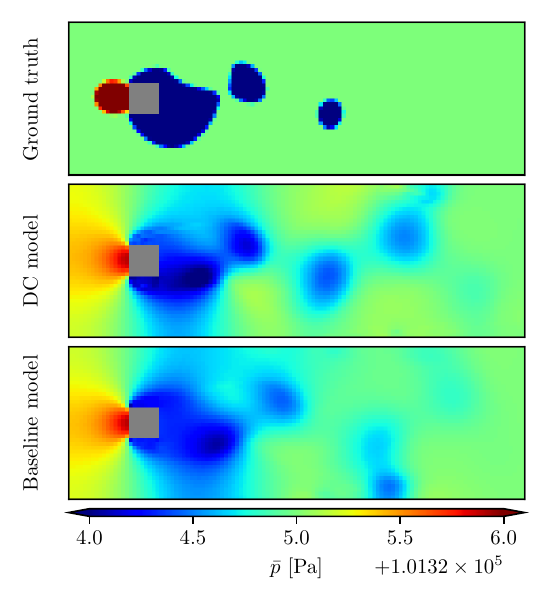}
	\caption{Comparison of pressure field at $t=20$ s between the ground truth, the Deep-Convection model, and the baseline solver.}
	\label{fig:mid_term_last_p}
\end{figure}

Although the improvement in the pressure modeling is less significative than in the rest of the variables, we observe a greater similitude between the ground truth vs. the field produced by the DC model than vs. the field of the baseline model, as Figure~\ref{fig:mid_term_last_p} shows. Another model limitation comes from the computation of hydrodynamic coefficients. Due to the problem caused by the law of the wall function (see Section~\ref{sec:bounding}), the model predicts drag coefficients with no enhancement compared with the baseline model for timesteps beyond those seen during training. For this reason, our model is inadequate for computing the temporal evolution of drag or lift coefficients. A solution to this problem could be to add an extra neural network to correct the cells affected by the law of the wall function, switch to a LES simulation, or impose the hydrodynamic coefficients as hard constraints in the loss function. This topic will be studied in upcoming works.

\begin{figure}[ht] 
	\centering
	\begin{subfigure}[b]{0.49\textwidth}
		\centering
		\subcaption{}
		\includegraphics[width=\textwidth]{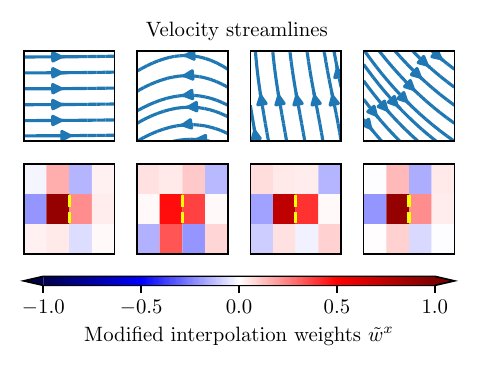}
	\end{subfigure}
	\begin{subfigure}[b]{0.49\textwidth}
		\centering
		\subcaption{}
		\includegraphics[width=\textwidth]{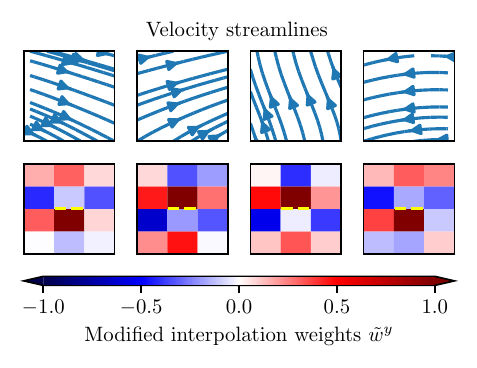}
	\end{subfigure}
	\caption{Modified interpolation coefficients generated by the neural network for a face $F$ (in dashed yellow line) and velocity streamlines. (a) Contribution to the x-velocity component. (b) Contribution to the y-velocity component.}
	\label{fig:interp_coeffs}
\end{figure}

Figure~\ref{fig:interp_coeffs} depicts the modified interpolation coefficients $\tilde{\textbf{w}}$ generated by the neural network. In the case of the x-components of these coefficients, we observe that the higher contribution corresponds to the upwind and downwind cells of the considered face. Hence, the main contribution of the model mimics the Upwind differencing scheme, taking into account the directionality of the flow while using the contribution of the rest of the cells to correct the predicted value. However, the distribution of the y-components is completely different. Although the higher positive coefficients belong to the upwind or the downwind cell, the contribution of the rest of the cells is more important than for the x-component case. This is due, as mentioned before, to the fact that the correction needed in the y-component is higher than in the x-component. Thus, although both velocity components follow the same physics, each needs a different degree of correction, which justifies using a separate encoder for each velocity component.

A crucial aspect that should be assessed is the computational cost associated with the prediction of the DC model compared with the baseline model, as well as the reduction in time compared to the fine simulation and the time invested in training the model. Table~\ref{tab:meshes} gathers the computation time per simulated timestep of the DC model and the baseline using the fine and the coarse meshes employing a single processor.
\begin{table}[ht]
	\centering
	\begin{tabular}{|c|c|}
		\hline
		\textbf{Model} 			& \textbf{Computation time per coarse timestep [s]} \\ \hline
		Baseline Fine      	    & 38                     \\ \hline
		Baseline Coarse         & 0.41                   \\ \hline
		Deep-Convection         & 0.61                  \\ \hline
	\end{tabular}
	\caption{Comparative computation time of the different models and discretizations.}
	\label{tab:meshes}
\end{table}

The difference in computation time between the baseline model using the coarse mesh and the DC model is due to the deep-learning framework. This time difference can be drastically reduced by employing the \textit{graph} TensorFlow mode instead of the \textit{eager} mode we employed, making it almost negligible. In our case, the \textit{eager} mode is justified in facilitating the development of the communication code between Python and C++. Despite this penalty, the DC model is 60 times faster than the fine baseline model. Thus, the results Figure~\ref{fig:mid_term_all} shows take about 4 minutes, with a 25\% error compared with the high-precision simulation, which takes 4 hours in a single core. The time for training the model (excluding the time for generating the training data) was 38.11 h, although the code can also be further optimized. Based on experiments in simpler cases, we estimate a reduction 35\% in training time could be possible by code optimization, added to the reduction due to the \textit{graph} TensorFlow mode. The time employed in converting the fine data to the coarse dimension and in constructing the dataset from the CFD simulation is negligible compared with the training time, being it in the order of seconds.

\subsection{Generalization}
Once we verified the model's good behavior under the training data, we then evaluate its generalization outside the training distribution. Theoretically, since the model exploits the local features of the physics, it can be applied to any flow with characteristics similar to those seen during training. For this purpose, we prepare two experiments: (1) extending the simulation time of the original case until $t=25$ s, and (2) applying the model to a case with two tandem columns with an extended domain.

\begin{figure}[H]
	\centering
	\includegraphics{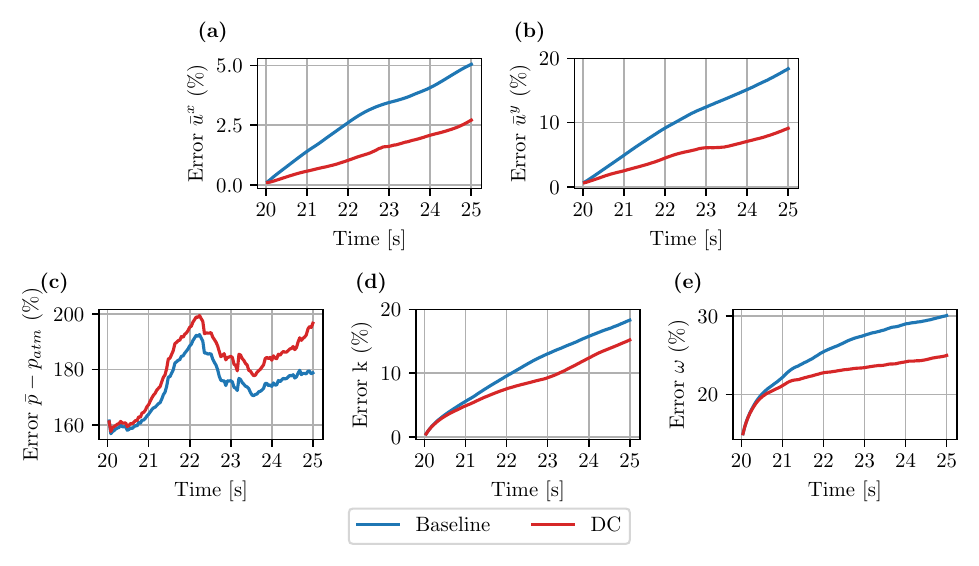}
	\caption{Comparison of results for the extended-in-time simulation between the baseline solver and the Deep Convection model. (a) Error on the x-component of the velocity. (b) Error on the y-component of the velocity. (c) Error on the pressure. (d) Error on the $k$ variable. (e) Error on the $\omega$ variable.}
	\label{fig:extended_results}
\end{figure}

In the first experiment, we follow a similar procedure to the previous section; both the Deep-Convection model and the baseline solver start from the same initial conditions at $t=20$ s, and they evolve 125 timesteps until reaching $t=25$ s. The ground truth data used for evaluating the performance of the models is obtained using the same configuration as for the training data. Figure~\ref{fig:extended_results} depicts the results of the experiment.

Figures~\ref{fig:extended_results}(a) and~\ref{fig:extended_results}(b) show that both components of the velocity follow a similar trend, reducing the DC model error by about 50\%. In the case of the pressure, both the DC and the baseline solver get a similar error, continuing the behavior of the training data. For the turbulent variables, Figures~\ref{fig:extended_results}(d) and~\ref{fig:extended_results}(e) show that the DC model reduces the error compared to the baseline solver, although in a minor grade than with the training data.

\begin{figure}[H]
	\centering
	\includegraphics{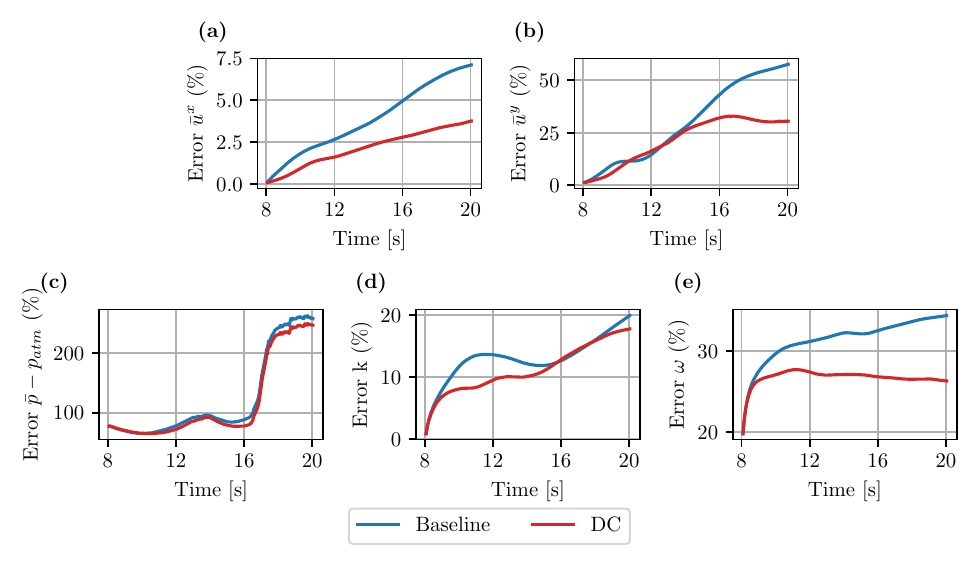}
	\caption{Comparison of results for the extended-in-domain and tandem simulation between the baseline solver and the Deep-Convection. (a) Error on the x-component of the velocity. (b) Error on the y-component of the velocity. (c) Error on the pressure. (d) Error on the $k$ variable. (e) Error on the $\omega$ variable.}
	\label{fig:tandem_results}
\end{figure}

In the second experiment, we extended the domain in the x-direction until $21D$, and we inserted a second column at a distance of $5D$ from the original, aligned and with the same size as the first one. Similarly to the training data, we discard the first initial timesteps due to the formation of the wake. However, since the domain is larger in this case, the wake formation takes longer, so we decided to exclude the first 8 s of the simulation. Thus, the experiment starts the simulation at $t=8$ s and lets both models evolve until $t=20$ s. Figure~\ref{fig:tandem_results} depicts the results of the experiment.

\begin{figure}[H]
	\centering
	\includegraphics{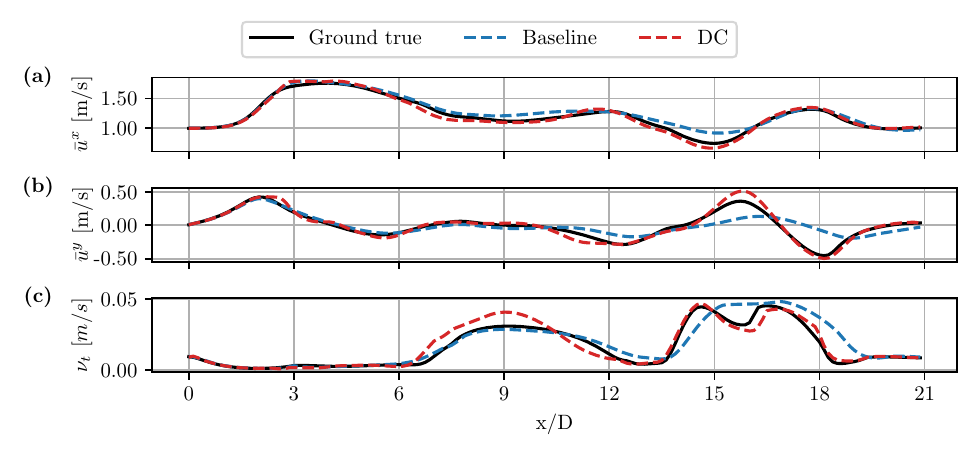}
	\caption{Comparison of the variables along the longitudinal line $y=3D/4$ at $t=20$ s. (a) Evolution of the x-velocity component. (b) Evolution of the y-velocity component. (c) Evolution of the turbulent viscosity $\nu_t$.}
	\label{fig:line_evolution}
\end{figure}

Although the tandem experiment´s flow regime is the same as in training data, the second column introduces new fluid behavior, becoming a more challenging experiment than just expanding the domain. Despite this difficulty, Figures~\ref{fig:tandem_results}(a) and~\ref{fig:tandem_results}(b) show that the Deep-Convection model reduces the error in both components of the velocity by around 50\% compared with the baseline. The reduction in the error is more evident after $t=14$ s, where the wake is fully developed in both columns, showing the same trend as with the training data. Thus, we expect a higher reduction in the error before $t=20$ s, both for the accumulation of the error in the baseline solver and for the full development of the flow behind the columns. Figure~\ref{fig:tandem_results}(c) shows agreement in the pressure with the previous results, with no significant variation compared with the baseline solver. The turbulent quantities follow a similar reduction than in the time-extended experiment, with a higher improvement expected in the following timesteps due to the total formation of the wakes. For a deeper exploration of the results, we include the evolution of some variables along the longitudinal line $y=3D/4$ (taking $y=0$ the lower wall at the bottom of the domain) at $t=20$ s in Figure~\ref{fig:line_evolution}. In particular, we observe a better agreement of the velocity generated by the DC model with the ground truth and a more precise concordance of the DC model to the complex evolution of the turbulence viscosity $\nu_t$, defined as the ratio between $k$ and $\omega$.

\begin{figure}[H]
	\centering
	\includegraphics{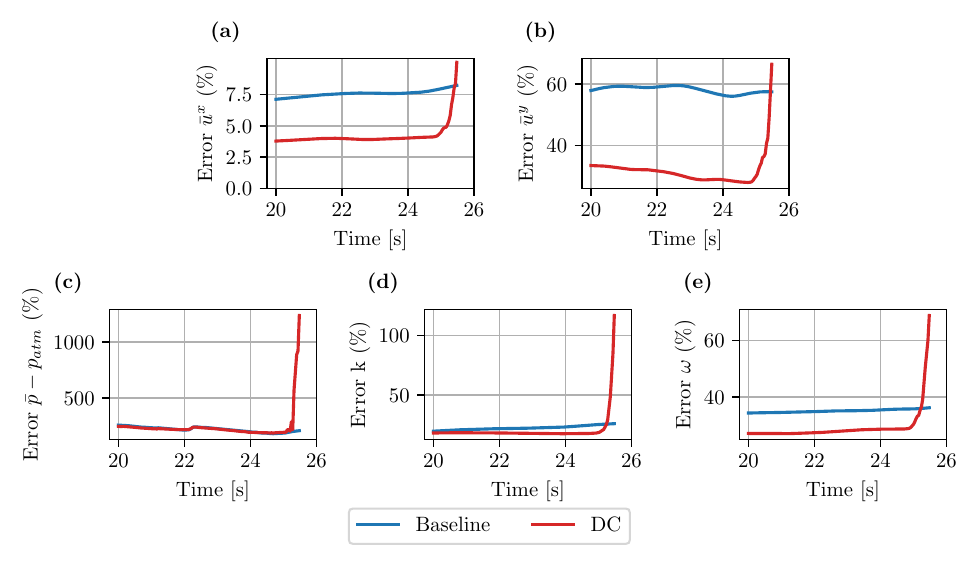}
	\caption{Comparison of results for the two-columns-in-tandem experiment extending the simulation time until failure. (a) Error on the x-component of the velocity. (b) Error on the y-component of the velocity. (c) Error on the pressure. (d) Error on the $k$ variable. (e) Error on the $\omega$ variable.}
	\label{fig:tandem_results_ext}
\end{figure}

These results show that the model accomplishes its objective of reducing the spatial discretization error in coarse meshes. Our model is unaware of the flow regime for two reasons: it receives the inputs in a normalized way and it does not see any information about the Reynolds number. Hence, it is unable to generalize to other Reynodls numbers different from the one it was trained for. This limitation is inherent to our model since it was not conceived for that purpose. Our objective was to apply our method to the fully turbulent regime, so while maintaining the Reynolds number of the simulation close to the Reynolds number of the training data, the model is expected to generalize well. In the case of extending its application to different Reynolds numbers, it would be necessary to include some knowledge of the grade of turbulence in the input of the neural network (e.g., turbulent viscosity of the Reynolds number). In the case of long-term simulations, the model is expected to produce degraded results due to the common problem of neural networks predicting time series. We extended the simulation of the two-columns-in-tandem in time to illustrate this behavior (see Figure~\ref{fig:tandem_results_ext}). A solution could be to monitor the residual of the NSE and switch to the baseline solver when the residual is over a certain threshold, as done in~\cite{ jeon-kim-vin}.

\section{Conclusions and Future Work}\label{conclusions}
We have developed and validated an OpenFOAM-embedded deep learning framework for incorporating fine-scale effects and flow structures into coarse mesh numerical simulations, minimizing the spatial discretization error at a reduced computational cost. We employed a feed-forward neural network to replace the traditional differencing scheme for the convective term. The model learns from simulations generated with a fine mesh and modifies the coarse solver to generate results close to the trained data. The framework is an end-to-end differentiable model, so we employed a discrete adjoint version of the CFD code to differentiate the physics automatically. We also developed a fast communication method between the programming languages of the programs (Python and C++). We trained the model with fine data projected to the coarse dimension for a flow past a square cylinder problem at a Reynolds number $Re=5\cdot10^5$, obtaining 60 times faster results with a 25\% error for both velocity components. We assessed the model's generalization outside the training distribution in a time-extended and a two-columns-in-tandem simulation. The model's response was similar in both cases, generating an error reduction of 50\% for both components of the velocity, a slight reduction in the turbulent variables, and a similar error in the pressure, compared with the baseline model. Hence, our model brings a new tool for generating precise results at a significantly reduced computational cost with application to a broad range of fluid dynamics problems.

As possible future lines of research, we highlight the inclusion of a second neural network to correct the pressures and the extension of the method to LES simulations, with the objective of improving the drag prediction. Another research line would be to extend the technique to unstructured meshes, considering new architectures like graph neural networks. Further research on the influence of the down-sampling operator would be highly desirable to understand the method's capabilities. Finally, following the recent developments in uncertainty quantification in fluid dynamics~\cite{ maulikTaira}, we find it promising to combine the presented methodology with machine-learning-based uncertainty quantification for both aleatoric (data)~\cite{ sunWang} and epistemic (models)~\cite{ morimoto} uncertainties.

\section*{Acknowledgement}
The authors have received funding from the Spanish Ministry of Science and Innovation projects with references TED2021-132783B-I00 funded by MCIN/ AEI /10.13039/5011000110\\33 and the European Union NextGeneration EU/ PRTR, PID2019-108111RB-I00 funded by MCIN/ AEI /10.13039/501100011033, PID2023-146678OB-I00 funded by MICIU/AEI/10.130\\39/501100011033 and by the European Union NextGenerationEU/ PRTR, the ``BCAM Severo Ochoa" accreditation of excellence CEX2021-001142-S / MICIN / AEI / 10.13039/5011\\00011033, PRE2020-093093 funded by MICIU/AEI/10.13039/501100011033 and EI ESF ``ESF Investing in your future”; the Spanish Ministry of Economic and Digital Transformation with Misiones Project IA4TES (MIA.2021.M04.008 / NextGenerationEU PRTR); and the Basque Government through the BERC 2022-2025 program, the Elkartek projects BEREZ-IA (KK-2023/00012) and EKIOCEAN (KK-2023/00097), and the Consolidated Research Group MATHMODE (IT1456-22) given by the Department of Education.

\section*{Declarations}
\textbf{Conflict of interests}. The authors have no conflict of interests to declare that are relevant to the content of this article.

\section*{Data Availability}
All the developments and results may be found and reproduced from the author´s Github repository (\url{https://github.com/jgonsie/Deep-Convection}).

\section*{Author Contributions}
JGS: Conceptualization, Methodology, Software, Data generation, Writing-original draft. DP: Methodology, Writing-review and editing. VN: Conceptualization, Writing-review and editing. VMC: Methodology, Writing-review and editing. MT: Software, Writing-review and editing.
\clearpage
\appendix
\section{Construction of training data from a CFD simulation}\label{generation_data}
After performing a CFD simulation with the features described in sections \ref{problem_description} and \ref{training_process}, we obtain $n$ snapshots, one for each timestep $t_i \in [t_0,t_1,...,t_n]$. We down-sample this fine data to the desired coarse resolution using the projection operator $\mathcal{P}_f^c$. Thus, we generate a projected version of the fine data that gathers each involved variable $(\bar{\textbf{u}}, \bar{p}, k, \omega)$ at time $t_i$, forming the training snapshots $d_i \in [d_0,d_1...,d_n]$. 

\begin{figure}[ht]
	\centering
	\includegraphics{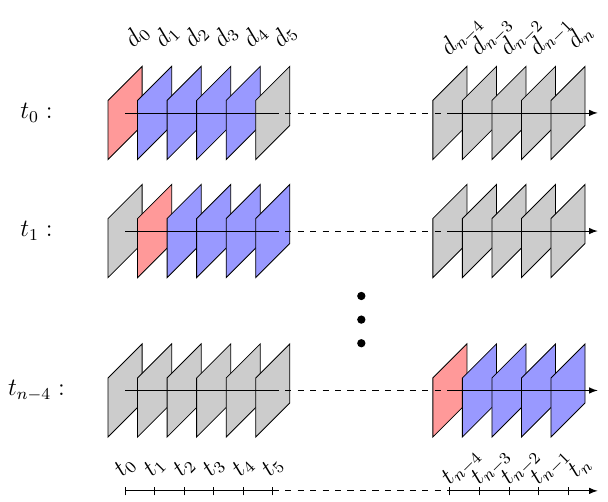}
	\caption{Grouping training snapshots $(d_i)$ for creating training data with $T=4$: red snapshot forms the input, blue snapshots are the output, and gray snapshots are not used for the correspondent $t_i$.}
	\label{fig:data_training}
\end{figure}

As seen in~\eqref{eq:loss}, the loss function is multi-step, which means that the loss is not computed regarding only the prediction of one timestep but regarding several. For this reason, we need to group the training snapshots in input-output samples as a function of the number of accumulated timesteps $T$. A higher number of accumulated timesteps will improve the generalization of our model, but it will produce a computationally more expensive training process. Thus, it is essential to select this variable properly. The grouping process, outlined in Figure~\ref{fig:data_training}, consists of going through all the available training snapshots and fixing $d_i$ as training input, while the following $T$ snapshots will be the training output. This process will produce $n-T$ training samples.

\clearpage

\bibliographystyle{elsarticle-num} 
\bibliography{bibliography}

\end{document}